\documentclass{article}
\usepackage{ijcai22}

\usepackage{times}
\usepackage{soul}
\usepackage{url}
\usepackage[hidelinks]{hyperref}
\usepackage[utf8]{inputenc}
\usepackage[small]{caption}
\usepackage{graphicx}
\usepackage{amsmath}
\usepackage{amsthm}
\usepackage{booktabs}
\usepackage{algorithmic}
\usepackage[linesnumbered,ruled,vlined]{algorithm2e}

\newcommand{\citep}[1]{\cite{#1}}
\newcommand{\citet}[1]{\citeauthor{#1} \shortcite{#1}}

\usepackage{amssymb}
\usepackage{tabu}
\usepackage{subcaption}
\usepackage{wrapfig}
\usepackage{multirow}
\usepackage{makecell}
\usepackage{csquotes}
\usepackage{mathtools}
\usepackage{booktabs}
\usepackage{hhline}
\usepackage{url}
\usepackage{bbm}

\usepackage[table,dvipsnames]{xcolor}         
\definecolor{myRed}{RGB}{229,101,101}
\definecolor{myBlue}{RGB}{101,101,229}
\definecolor{citegreen}{RGB}{0,180,0}
\definecolor{linkblue}{RGB}{0,0,180}
\definecolor{myGray}{gray}{0.92}

\let\httpsurl\url

\usepackage{todonotes}

\definecolor{darkblue}{RGB}{0,0,210}

\newcommand{\ub}[1]{\textbf{#1}}
\newcommand{\cm}{\checkmark}

\usepackage{amsmath}
\DeclareMathOperator*{\argmax}{argmax}

\newcommand{\Dpixel}{\mathcal{D}_\mathit{pixel}}
\newcommand{\Dimage}{\mathcal{D}_\mathit{image}}
\let\labelname\texttt

\usepackage{xspace}
\makeatletter
\DeclareRobustCommand\onedot{\futurelet\@let@token\@onedot}
\def\@onedot{\ifx\@let@token.\else.\null\fi\xspace}
\makeatother

\def\eg{{e.g}\onedot} 
\def\ie{{i.e}\onedot} 
\def\vs{{vs}\onedot}

\usepackage{tabularx}

\newlength\myindent
\usepackage[capitalize,noabbrev]{cleveref}

\SetKwInput{KwInput}{Input}
\SetKwInput{KwOutput}{Output}
\SetKwInput{KwRequire}{Require}

\title{One Weird Trick to Improve Your Semi-Weakly Supervised Semantic Segmentation Model}

\author{
Wonho Bae$^1$
\and
Junhyug Noh$^2$\and
Milad Jalali Asadabadi$^{1}$\And
Danica J.\ Sutherland$^{1,3}$
\affiliations
$^1$University of British Columbia\\
$^2$Lawrence Livermore National Laboratory
$^3$Alberta Machine Intelligence Institute
\emails
\{whbae, miladj7, dsuth\}@cs.ubc.ca,
noh1@llnl.gov
}

\begin{document}

\maketitle

\begin{abstract}
  Semi-weakly supervised semantic segmentation (SWSSS) aims to train a model to identify objects in images based on a small number of images with pixel-level labels, and many more images with only image-level labels. 
  Most existing SWSSS algorithms extract pixel-level pseudo-labels from an image classifier -- a very difficult task to do well,
  hence requiring complicated architectures and extensive hyperparameter tuning on fully-supervised validation sets.
  We propose a method called \emph{prediction filtering}, which instead of extracting pseudo-labels, just uses the classifier as a classifier:
  it ignores any segmentation predictions from classes which the classifier is confident are not present.
  Adding this simple post-processing method to baselines gives results competitive with or better than prior SWSSS algorithms.
  Moreover, it is compatible with pseudo-label methods: adding prediction filtering to existing SWSSS algorithms further improves segmentation performance.
\end{abstract}

\section{Introduction} \label{sec:introduction}
Recent semantic segmentation algorithms \cite{DeeplabV3+:2018Chen,Wider:2019Wu} have successfully solved challenging benchmark datasets for semantic segmentation tasks like PASCAL VOC~\citep{VOC:Everingham2015} and MS COCO~\citep{COCO:Lin2014}.
To do so, however, they use a large number of pixel-level annotations,
which require extensive human labor to obtain.
Great attention in computer vision research has thus turned to \emph{weakly-supervised} learning \citep{OAA:2019Jiang,SEAM:2020Wang,MCIS:2020Sun}.
Weakly-supervised semantic segmentation aims to classify each pixel of test images, trained only on image-level labels
(whether a class is present in the image, but not its location).
Although weakly-supervised approaches have seen success in both semantic segmentation and object localization tasks, \citet{EvalWSOL:Choe2020} cast significant doubt on their validity and practicality.
They argue that although weakly-supervised learning algorithms are designed to be trained only on image-level labels, they inevitably use explicit pixel-level labels (or, equivalently, manual judgement of outputs) in hyperparameter tuning.
Since at least \emph{some} fully-supervised inputs are necessary,
\citeauthor{EvalWSOL:Choe2020} point out that simply using a small number of these,
\eg five per class,
to train a fully-supervised localization model
substantially outperforms a weakly-supervised counterpart.
To still take advantage of less-expensive weakly-supervised data points, though,
perhaps the most natural change is
to the semi-weakly supervised semantic segmentation (SWSSS) task:
here only a small number of pixel-level labels are provided, as well as a large number of image-level labels.

Segmentation networks trained on a small number of pixel-level labels often confuse similar classes, \eg \labelname{cat} and \labelname{horse}, as they architecturally tend to focus on local features rather than farther-away distinguishing areas.
Thus, the additional supervision from image-level labels can be potentially quite helpful.
Most existing SWSSS methods generate pseudo-labels from a classifier using class activation maps (CAMs) \cite{CAM:Zhou2015},
then train a segmentation network using both pseudo-labels and true pixel labels.
These pseudo-labels, however, are difficult to extract:
they tend to focus on small discriminative regions of objects,
ignoring less-distinctive bulks of objects and often including nearby pixels that are not part of objects.
Our analysis shows that as the baseline segmentation model improves with more training data,
the pseudo-labels quickly provide more incorrect than correct supervision to what the model already would have predicted.
Previous methods have thus employed additional information, such as saliency maps, or additional processing methods, adding complexity and many more hyperparameters to tune on a fully-supervised validation set.
We use weak supervision data differently,
without requiring any side information and introducing far fewer new hyperparameters.

\begin{figure}[t!]
  \centering
  \includegraphics[width=0.48\textwidth]{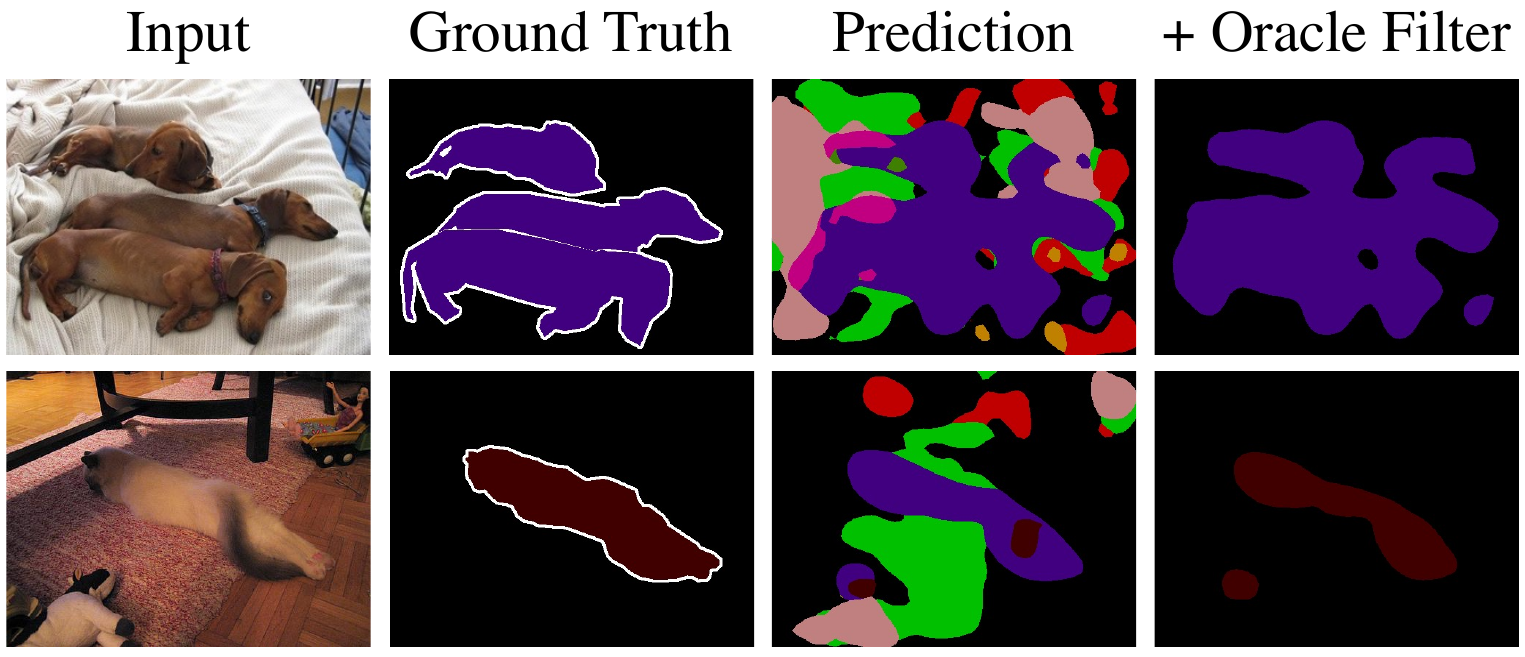}
  \caption{Filtering this model's predicted classes drastically improves segmentation quality.}
  \label{fig:gt_filtering}
  \vspace*{-3.5mm}
\end{figure}

%
%
To motivate our method, consider \cref{fig:gt_filtering}.
Baseline models with small training sets predict many classes which are not present in the image.
If we ignore the predictions for the class of any pixel which are not present in the image at all, which we term \textit{oracle filtering}, then the segmentation performance improves dramatically.
Inspired by this, we propose a simple algorithm we call \textit{prediction filtering}.
Prediction filtering uses a multi-label classifier trained on only image-level labels
to filter out segmentation predictions deemed very unlikely by the classifier, replacing predictions for those pixels with the next-most-likely class allowed through the filter.
It is compatible with any segmentation model, and
the threshold for ``very unlikely'' is the only new hyperparameter introduced.

Although the classifier is not perfect, because it is trained on a large weakly-supervised set, its predictions tend to be quite accurate.
Moreover, it is trying to solve an easier problem than the segmentation network, using a different architecture.
As we will see in the experiments, even without any additional weakly-supervised data, prediction filtering tends to improve the segmentation performance.
When applied to baseline segmentation models, the performance significantly improves; adding it to the baseline model variant from \citet{Dual:2020Luo} achieves (to our knowledge) the new highest performance on PASCAL VOC in the SWSSS regime.
As prediction filtering is so general, it can even be easily applied to models which already exploit weakly-supervised data via pseudo-labels;
doing so on the state-of-the-art SWSSS algorithms \citep{CCT:2020Ouali,Dual:2020Luo} yields a new model with significantly higher performance,
with more improvement for models trained on fewer fully-labeled images.

\section{Approaches to Semantic Segmentation}
\label{sec:related_work}

\paragraph{Fully-supervised semantic segmentation.}
In this task, we have a training set of images with pixel-level class labels:
$\Dpixel = \{(x_i, y_i)\}_{i=1}^M$, where
$x_i \in \mathbb{R}^{3 \times H_i\times W_i}$ are images
and $y_i \in \{0,1\}^{K\times H_i\times W_i}$ are pixel labels,
with $K$ the number of classes.
Our goal is to find a model that can predict pixel labels $y$ given a new image $x$.

Current approaches are mostly based on convolutional networks.
One important factor to their success is using larger receptive fields via dilated convolutional layers \citep{Dilated:2015Yu,DeeplabV3:2017Chen}.
Even so, state-of-the-art algorithms still misclassify many pixels
when multi-label classifiers on the same data obtain near-perfect accuracy.
We conjecture this is because segmentation models still miss global structure of the image when looking at an individual pixel.
We will exploit that a classifier ``looks at images differently.''

\paragraph{Weakly-supervised semantic segmentation.}
To avoid extensive human labor for pixel-level annotation, there have been many attempts to replace pixel-level labels with image-level labels:
$\Dimage = \{(x_i, z_i)\}_{i=1}^N$,
with $z_i \in \{0,1\}^{K}$ a ``logical or'' of each channel of the unknown $y_i$.
We still want a model to produce $y$.
The most common pipeline for weakly-supervised semantic segmentation is to generate a class activation map (CAM) \citep{CAM:Zhou2015}, refine it with various post-processing methods, then use it as a pseudo-label to train a semantic segmentation network.
However, CAMs tend to focus only on discriminative regions of an object.
Prior work has attempted to expand the CAM to entire objects by masking out parts of an image \citep{HaS:Singh2017} or intermediate feature \citep{ACoL:Zhang2018,ADL:Choe2019};
these methods do indeed expand the resulting CAM, but often too much, in very unstable ways.
Another popular approach to grow the CAM regions, via methods proposed by \citet{CRF:Krahenbuhl2011}, \citet{DSRG:2018Huang}, or \citet{PSA:2018Ahn}, until it converges to the object region \cite{Boundary:2020Chen}.


\paragraph{Semi-weakly supervised semantic segmentation.}
\citet{EvalWSOL:Choe2020} point out fundamental problems with weakly-supervised learning, as discussed in \cref{sec:introduction}.
We thus consider combining a small number of pixel-annotated images $\Dpixel$ with many weakly-supervised images $\Dimage$, which we refer to as semi-weakly supervised semantic segmentation (SWSSS).
Although they have not used exactly this name, many papers have already addressed this setting.
Broadly speaking, the most common approach is to generate pseudo-labels from a CAM, and use these in combination with true labels to train a segmentation network
\citep{WSSL:2015Papandreou,MDC:2018Wei,GAIN:2018Li,CCT:2020Ouali,Dual:2020Luo,AdvCAM:2021Lee,vessel2022dang,semi_context_const2021lai}.
(For lack of space, we unfortunately elide all details of these approaches.)
Because image-level labels have no spatial information, however, it is fundamentally difficult to make accurate pseudo-labels.
As we will now argue, as the number of pixel labels increases and base models improve, the benefit of pseudo-labels drastically diminishes.

\label{sec:problem}



\begin{figure*}[t]
    \centering
    \includegraphics[width=0.8\textwidth]{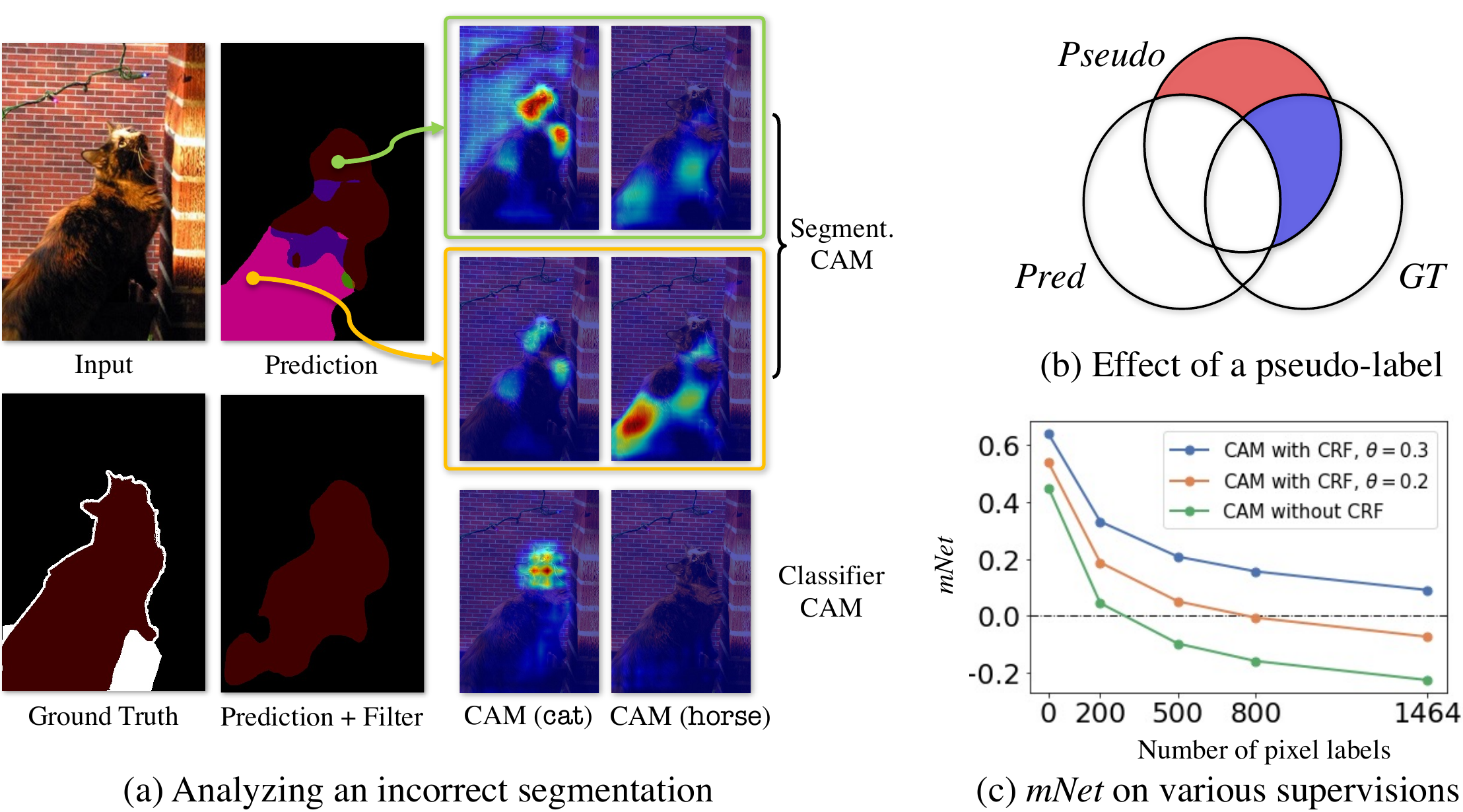}
    \caption{A segmentation network trained on few pixel-level labels confuses similar classes, as it does not capture the global features (panel a). To alleviate this issue, pseudo-labels have been widely used; this approach is sensitive to the values of additional hyperparameters, and the quality of its supervision drastically diminishes as the number of pixel labels increases (panel c).}
    \vspace*{-2mm}
    \label{fig:problems}
\end{figure*}


To demonstrate this, we train a DeeplabV1 segmentation network \citep{DeeplabV1:2015Chen} on $\Dpixel$ consisting of $1{,}464$ images from PASCAL VOC training set, and a VGG16 classifier on $\Dimage$ containing all $10{,}582$ images in the full training set.
To examine what part of the image causes this prediction, we extract CAMs for \labelname{cat} and \labelname{horse} classes using GradCAM~\cite{Gradcam:Selvaraju2016}.
Although the classifier confidently predicts that only the \labelname{cat} class is present in the given image ($\Pr(\labelname{cat}) = 0.97$, $\Pr(\labelname{horse}) = 0.03$), \cref{fig:problems}(a) shows that the segmentation model predicts most of the cat's body as \labelname{horse} (pink region).
The CAM shows that the classifier makes this decision based on the most discriminative region of the object, \ie the cat's head.
The segmentation model does the same at the green (top) location, correctly predicted as \labelname{cat};
at the yellow (middle) location, however, the \labelname{horse} prediction is based mostly on the more ambiguous body area.
As $\Dpixel$ grows, this phenomenon is largely alleviated; it seems $1{,}464$ images were not enough for the segmentation model to learn ``where to look.''


Supervision added by previous models from classifier CAMs, then,
will also tend to focus on discriminative regions of an object,
and can therefore be misleading.
To estimate the effectiveness of the pseudo-labels, we define a measure called \textit{mNet}, the mean over classes $c$ of
\[
    \mathit{net}_c =
    \frac{\sum_{i=1}^N \!\!\splitfrac{%
        \textcolor{myBlue}{\operatorname{area}\left(
            \left( \mathit{Pseudo}_{i,c} \setminus \mathit{Pred}_{i,c} \right)
            \cap \mathit{GT}_{i,c}
        \right)} }{%
        - \textcolor{myRed}{\operatorname{area}\left(
            \left( \mathit{Pseudo}_{i,c} \setminus \mathit{Pred}_{i,c} \right)
            \setminus \mathit{GT}_{i,c}
        \right))} }
    }{
        \sum_{i=1}^N \operatorname{area}(\mathit{Pseudo}_{i,c} \setminus \mathit{Pred}_{i,c})
    }
    \label{eq:net}
.\]
Here
the subscript $\cdot_{i,c}$ refers to the set of pixels of $i$-th training image whose label is $c$;
$\mathit{GT}$ refers to the ground truth labels $y$,
$\mathit{Pred}$ to the predicted labels from a baseline segmentation model,
and $\mathit{Pseudo}$ to the CAM-based pseudo-labels.
The first (blue) term in the numerator measures how much correct supervision the pseudo-label adds (see \cref{fig:problems}(b));
the second (red) term, how much incorrect supervision is added.
The denominator combines these two regions.
$\mathit{mNet}$ does not exactly measure how much the pseudo-labels help predictions, but it gives a rough sense of how correct their information is.

\Cref{fig:problems}(c) shows that ``plain'' CAM (in green, lowest) indeed helps when $\Dpixel$ is very small, but as it grows, $\mathit{mNet}$ even becomes negative.
It is possible to improve these predictions by, for instance, post-processing with a CRF (blue line, top).
This, however, requires a complicated structure with several additional hyperparameters to tune on a fully-supervised validation set;
setting these parameters correctly significantly affects performance,
as shown \eg by the substantially worse $\mathit{mNet}$ when changing the threshold for a foreground region $\theta$ from $0.3$ to $0.2$ (orange line, middle).

\section{Prediction Filtering}
\label{sec:prediction_filtering}
\paragraph{Motivation.}
Given a segmentation network $f$,
whose output on the $i$-th image is $f(x_i) \in \mathbb{R}^{K \times H\times W}$,
the final prediction at each pixel $(h, w)$ is normally
\begin{align}
    \hat{y}_{h,w} = \argmax_{c \in \mathcal K} {f(x_i)_{c, h, w}}
    \label{eq:normal}
,\end{align}
where $\mathcal K = \{1, 2, \dots, K\}$.
\emph{Oracle filtering} (\cref{fig:gt_filtering}) instead only considers classes actually present in the image, maximizing over $\mathcal{\tilde K}_i = \{ c : z_{i,c} = 1 \}$.
This improves segmentation performance substantially;
the mIoU (mean intersection over union, the standard segmentation performance metric) 
of a DeeplabV1 segmentation network trained on $\Dpixel$ with $1{,}464$ images improves from $61.8$ to $70.6$.
We conjecture this is because the segmentation network has not learned to appropriately consider global features when predicting at each pixel of the image,
while the classifier, solving an easier problem with more data, can immediately identify relevant areas.

\begin{table*}[t!]
\centering
\fontsize{9.0}{12.0}\selectfont
\begin{tabular}{c|c|l|c|c|c|l}
\hline
Backbone & Add. 9.1K Images & \multicolumn{1}{c|}{Method} & Bkg. Cues & CRF & Pred. Filter & $\!$mIoU \\
\hline\hline
\multirow{8}{*}{VGG} 
  & -- & DeeplabV1~\citep{DeeplabV1:2015Chen}  & -- & \cm & -- &  61.8 \\
  \hhline{*{1}{~}*{6}{|-}}
  & \multirow{6}{*}{Image-level} & \cellcolor{myGray}DeeplabV1~\citep{DeeplabV1:2015Chen}  & \cellcolor{myGray}-- & \cellcolor{myGray}\cm & \cellcolor{myGray}\cm & \cellcolor{myGray}\ub{67.4} \\
  & & WSSL~\citep{WSSL:2015Papandreou}  & --  & \cm & -- & 64.6 \\
  & & \cellcolor{myGray}WSSL~\citep{WSSL:2015Papandreou} & \cellcolor{myGray}-- & \cellcolor{myGray}\cm & \cellcolor{myGray}\cm & \cellcolor{myGray}67.1 \\
  & & \citet{SemiGAN2017Souly}  & -- & -- & -- & 65.8$^*$ \\
  & & MDC~\citep{MDC:2018Wei}  & \cm   & \cm & -- & 65.7$^*$  \\
  & & FickleNet~\citep{FickleNet:2019Lee} & \cm  & \cm & -- & 65.8$^*$ \\ 
  \hhline{*{1}{~}*{6}{|-}}
  & Pixel-level & DeeplabV1~\citep{DeeplabV1:2015Chen}  & -- & \cm & -- & 69.0  \\
\Xhline{2\arrayrulewidth}
\multirow{4}{*}{VGG-W}
  & -- & DeeplabV1-W~\citep{Dual:2020Luo}  & -- & \cm & -- & 69.2 \\
  \hhline{*{1}{~}*{6}{|-}}
  & \multirow{3}{*}{Image-level} & \cellcolor{myGray}DeeplabV1-W~\citep{Dual:2020Luo}  & \cellcolor{myGray}-- & \cellcolor{myGray}\cm & \cellcolor{myGray}\cm & \cellcolor{myGray}73.8 \\
  & & DualNet~\citep{Dual:2020Luo} & \cm  & \cm & -- & 73.9 \\
  & & \cellcolor{myGray}DualNet~\citep{Dual:2020Luo} & \cellcolor{myGray}\cm & \cellcolor{myGray}\cm & \cellcolor{myGray}\cm & \cellcolor{myGray}\ub{75.1} \\ \Xhline{2\arrayrulewidth}
\multirow{9}{*}{ResNet}
  & \multirow{2}{*}{--} & DeeplabV3~\citep{DeeplabV3:2017Chen} & -- & \cm & -- & 72.4 \\
  & & \citet{semi_context_const2021lai} & -- & -- & -- & 74.5 \\
  \hhline{*{1}{~}*{6}{|-}}
  & \multirow{6}{*}{Image-level} & \cellcolor{myGray}DeeplabV3~\citep{DeeplabV3:2017Chen} & \cellcolor{myGray}-- & \cellcolor{myGray}\cm & \cellcolor{myGray}\cm & \cellcolor{myGray}75.3 \\ 
  & & \citet{semi_context_const2021lai} & -- & -- & -- & 76.1 \\
  & & CCT~\citep{CCT:2020Ouali} & -- & \cm & -- & 74.7 \\
  & & \cellcolor{myGray}CCT~\citep{CCT:2020Ouali} & \cellcolor{myGray}-- & \cellcolor{myGray}\cm & \cellcolor{myGray}\cm & \cellcolor{myGray}76.0 \\
  & & AdvCAM~\citep{AdvCAM:2021Lee} & -- & \cm & -- & 76.1 \\
  & & \cellcolor{myGray}AdvCAM~\citep{AdvCAM:2021Lee} & \cellcolor{myGray}-- & \cellcolor{myGray}\cm & \cellcolor{myGray}\cm & \cellcolor{myGray}\ub{77.1} \\ 
  \hhline{*{1}{~}*{6}{|-}}
  & Pixel-level & DeeplabV3~\citep{DeeplabV3:2017Chen} & -- & \cm &-- & 77.4 \\ 
\Xhline{2\arrayrulewidth}
\multirow{4}{*}{ResNet-W}
  & -- & DeeplabV3-W~\citep{Dual:2020Luo} & -- & \cm & -- & 76.2 \\ 
  \hhline{*{1}{~}*{6}{|-}}
  & \multirow{3}{*}{Image-level} & \cellcolor{myGray}DeeplabV3-W~\citep{Dual:2020Luo} & \cellcolor{myGray}-- & \cellcolor{myGray}\cm & \cellcolor{myGray}\cm & \cellcolor{myGray}\ub{77.5} \\ 
  & & DualNet~\citep{Dual:2020Luo} & \cm  & \cm & -- & 76.7 \\
  & & \cellcolor{myGray}DualNet~\citep{Dual:2020Luo} & \cellcolor{myGray}\cm  & \cellcolor{myGray}\cm & \cellcolor{myGray}\cm & \cellcolor{myGray}77.3 \\
\Xhline{2\arrayrulewidth}
\multirow{3}{*}{HRNetV2-W48}
  & -- & OCRNet~\citep{OCRNet:2021Strudel} & -- & -- & -- & 74.0 \\ 
  \hhline{*{1}{~}*{6}{|-}}
  & Image-level & \cellcolor{myGray}OCRNet~\citep{OCRNet:2021Strudel}& \cellcolor{myGray}-- & \cellcolor{myGray}-- & \cellcolor{myGray}\cm & \cellcolor{myGray}\ub{75.8} \\ 
  \hhline{*{1}{~}*{6}{|-}}
  & Pixel-level & OCRNet~\citep{OCRNet:2021Strudel}& -- & -- & -- & 77.7 \\ 
\Xhline{2\arrayrulewidth}
\end{tabular}
\vspace*{-1.0mm}
\caption{Comparison of the state-of-the-art methods on $1{,}464$ images with pixel labels. The ``Add. 9.1K Images'' column gives which type of supervision is used for the 9.1K additional images (augmented dataset). Numbers marked with $*$ are as reported by the corresponding paper.}
\label{tbl:sota}
\vspace*{-1.0mm}
\end{table*}

\newcolumntype{C}{>{\centering\arraybackslash}p{3.2em}}
\newcolumntype{D}{>{\centering\arraybackslash}p{6.1em}}
\begin{table*}[t!]
\setlength{\tabcolsep}{0.7pt}
\centering
\fontsize{6.5}{9.0}\selectfont
\begin{tabu}{D|CCCCCCCCCCCCCCCCCCCC|C} 
\hline
Method & plane & bike & bird & boat & bottle & bus & car & cat & chair & cow & table & dog & horse & mbike & person & plant & sheep & sofa & train & tv & mIoU \\
\hline\hline
DeeplabV1 & 71.1 & 37.1 & 78.5 & 52.7 & 58.3 & 79.4 & 72.0 & 73.2 & 20.6 & 58.0 & 56.1 & 66.5 & 55.9 & 76.1 & 76.4 & 40.6 & 65.4 & 42.8 & 65.2 & 53.3 & 61.4 \\
+ Pred.\ Filter    & \ub{76.2} & \ub{38.9} & \ub{82.2} & \ub{58.2} & \ub{61.2} & \ub{85.1} & \ub{76.8} & \ub{84.4} & \ub{22.6} & \ub{73.0} & \ub{56.2} & \ub{79.3} & \ub{76.2} & \ub{82.4} & \ub{78.2} & \ub{46.0} & \ub{80.4} & \ub{43.6} & \ub{71.5} & \ub{55.2} & \ub{67.6} \\
\hline
DeeplabV3-W & \ub{89.3} & 60.2 & 80.5 & 56.4 & 73.7 & 92.5 & 83.8 & \ub{92.4} & 31.1 & 83.6 & \ub{69.9} & \ub{85.3} & 81.9 & 84.8 & \ub{85.4} & 63.2 & 84.2 & 52.7 & \ub{83.9} & \ub{69.8} & 76.1 \\
+ Pred.\ Filter  & \ub{89.3} & \ub{61.7} & \ub{81.5} & \ub{58.0} & \ub{73.8} & \ub{92.6} & \ub{84.3} & 91.3 & \ub{34.4} & \ub{84.9} & 69.8 & 85.1 & \ub{89.4} & \ub{86.2} & 85.1 & \ub{64.3} & \ub{89.0} & \ub{54.3} & 83.2 & 68.1 & \ub{77.2} \\
\Xhline{2\arrayrulewidth}
\end{tabu}
\vspace*{-1.0mm}
\caption{Evaluation results of DeeplabV1 (VGG-based) and DeeplabV3-W (ResNet-based) models on the test set.}
\vspace*{-4.0mm}
\label{tbl:test}
\end{table*}

\paragraph{Prediction filtering.}
Inspired by this phenomenon, we propose a simple post-processing method, \emph{prediction filtering}.
Given $D_{image}$ with a large number of images, we can train a highly accurate multi-label classifier $g$;
a ResNet50 achieves $99\%$ accuracy and $97.5\%$ average precision on PASCAL VOC.
Hence, we constrain predictions to come from the classifier's predictions instead of ground truth classes,
$\mathcal{\hat K}_i = \{ c : g(x_i)_c > \tau \}$,
where $g(x_i)_{c}$ is the output logit of $g$ for class $c$,
and $\tau$ is a threshold to determine the presence of a class in an image.
We provide full pseudocode in the appendix.


Compared to other SWSSS algorithms, prediction filtering has several advantages.
First, the architecture is simple. It only requires an additional classifier, which can be trained in parallel with the segmentation network; most existing methods require training a classifier first.
Second, it requires only a single additional hyperparameter, the threshold $\tau$, far fewer than required by other SWSSS algorithms.
For instance, MDC~\cite{MDC:2018Wei} requires two thresholds to determine the foreground and background regions, in addition to selecting the number of dilation layers with different rate for each layer.
(We provide a comprehensive comparison of hyperparameter counts in the appendix).
Prediction filtering thus minimizes the requirements on the additional fully-supervised validation set.
Third, it can be independently added to any segmentation algorithm, including existing SWSSS algorithms; we do so in our experiments. 

\paragraph{Effect on performance.}

Prediction filtering helps performance when an incorrect prediction is filtered out and the ``backup'' prediction is correct;
it hurts when a correctly-predicted object is incorrectly filtered.
It can also change an incorrect prediction to a different incorrect prediction;
this can in fact either increase or decrease the mIoU score.

For a non-perfect classifier and reasonable setting of $\tau$, it is conceivable for prediction filtering to hurt segmentation performance -- although we did not observe this in our experiments.
There is always a value of the threshold $\tau$, however, for which prediction filtering at least does not hurt:
just take $\tau \to -\infty$, in which case no predictions are changed.
As $\tau$ increases, it likely (though not certainly) removes incorrect predictions before it begins removing any correct predictions.
In another extreme, for a perfect classifier, prediction filtering approaches oracle filtering;
clearly, oracle filtering may not achieve perfect performance, but it can only help.


\begin{figure*}[t!]
    \centering
    \includegraphics[width=0.95\textwidth]{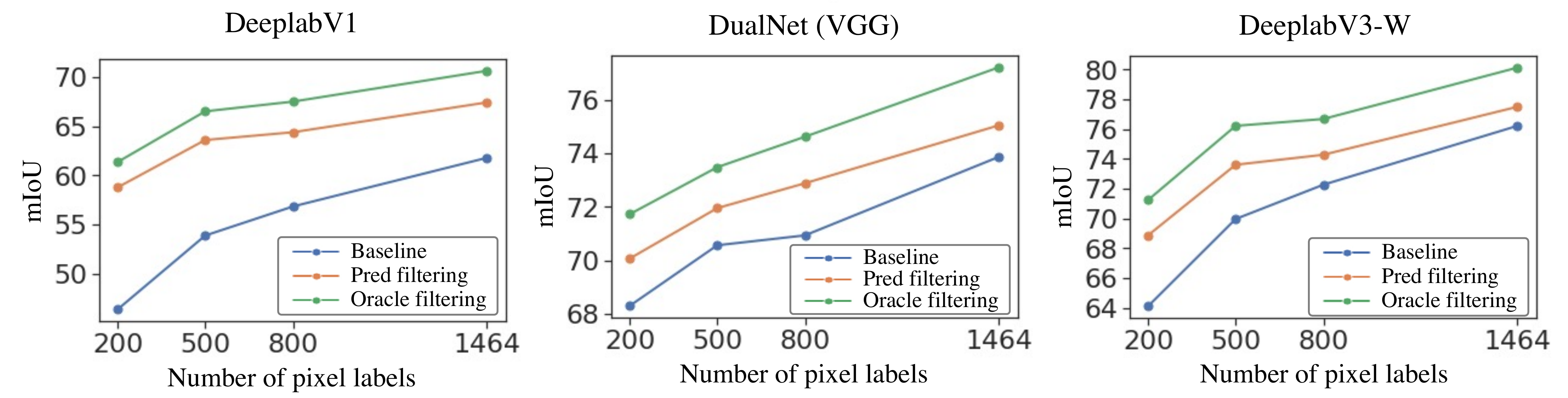}
    \caption{Performance of the prediction filtering on various models and levels of supervision (M).}
    \label{fig:various_supervision}
\end{figure*}

\newcolumntype{F}{>{\centering\arraybackslash}p{2.0em}}
\newcolumntype{E}{>{\centering\arraybackslash}p{4.0em}}
\begin{table*}[t!]
\begin{minipage}{.53\textwidth}
\centering
\fontsize{9.0}{12.0}\selectfont
\begin{tabu}{cc|EE|EE}
\hline
\multirow{2}{*}{Filtering} &
\multirow{2}{*}{CRF} &
\multicolumn{2}{c|}{DeeplabV1} &
\multicolumn{2}{c}{DeeplabV3}\\
\cline{3-6}
& & $M$=$500$ & $M$=$1{,}464$ & $M$=$500$ & $M$=$1{,}464$ \\
\hline
\hline
 & & 49.6 & 57.2 & 57.0 & 70.6 \\
 \cm & & 61.4 & 64.6 & 64.8 & 74.0  \\
 & \cm & 53.9 & 61.8 & 58.4 & 72.4 \\
 \cm & \cm & \ub{63.6} & \ub{67.4} & \ub{65.9} & \ub{75.3} \\
\Xhline{2\arrayrulewidth}
\end{tabu}
\end{minipage}%
\begin{minipage}{.52\textwidth}
\centering
\fontsize{9.0}{12.0}\selectfont
\begin{tabu}{cc|FFFF}
\hline
\multirow{2}{*}{Filtering} & \multirow{2}{*}{CRF} & \multicolumn{4}{c}{Pixel-level labels ($M$)} \\
\cline{3-6}
 & & 200 & 500 & 800 & 1,464 \\
\hline
\hline
 & & 43.3 & 49.9 & 52.7 & 57.2 \\
 \cm & & 46.1 & 54.0 & 57.1 & 62.0\\
 & \cm & 46.3 & 53.9 & 56.8 & 61.8 \\
 \cm & \cm & \ub{48.2} & \ub{56.5} & \ub{59.8} & \ub{64.9} \\
\Xhline{2\arrayrulewidth}
\end{tabu}
\end{minipage}%
\caption{{Left}: Performance of model variants with $\lvert\Dpixel\rvert = M$ and $\lvert\Dimage\rvert = 10,582$. {Right}: the same, for a DeeplabV1 baseline, but with the classifier trained only on the same $M$ images in $D_{pixel}$.}
\vspace*{-1.0mm}
\label{tbl:crf_and_partial}
\end{table*}

\section{Experimental Evaluation}
\label{sec:experiment}


\paragraph{Dataset.}

We evaluate prediction filtering on PASCAL VOC 2012 \citep{VOC:Everingham2015} 
which contains $10{,}582$ training, $1{,}449$ validation, and $1{,}456$ test images.
For SWSSS, we follow the training splits of \citet{CCT:2020Ouali}, where $1{,}464$ images are used for $\Dpixel$.
As with previous work, we evaluate segmentation performance by mean Intersection over Union (mIoU),
generally on the validation set.
Test set performance is obtained from the PASCAL VOC evaluation server, without any tricks such as multi-scale or flipping.

\paragraph{Implementation.}
To verify the robustness of our method, we experiment with five semantic segmentation baselines: DeeplabV1 (based on VGG16), DeeplabV3 (based on ResNet-101), their deeper variants with wider receptive fields used by \citet{Dual:2020Luo} which we call DeeplabV1-W and DeeplabV3-W, and a Transformer model called OCRNet~\cite{OCRNet:2021Strudel} (based on HRNetV2-W48~\cite{HRNet2020Wang}).
For prediction filtering, we use a ResNet50 classifier.
We also apply prediction filtering to several existing SWSSS models.
Although CRF post-processing is no longer commonly used in semantic segmentation tasks,
in our experiments it still significantly improves the performance of models trained on a small number of pixel-level labels.
We thus apply CRFs as default except when otherwise specified.




\begin{figure*}[t!]
    \centering
    \begin{subfigure}[b]{0.47\textwidth}
        \includegraphics[width=\textwidth]{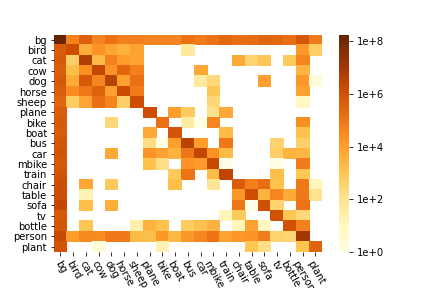}
        \caption{Baseline model}
        \label{fig:confusion-nofilter}
    \end{subfigure}
    \hfill
    \begin{subfigure}[b]{0.47\textwidth}
        \includegraphics[width=\textwidth]{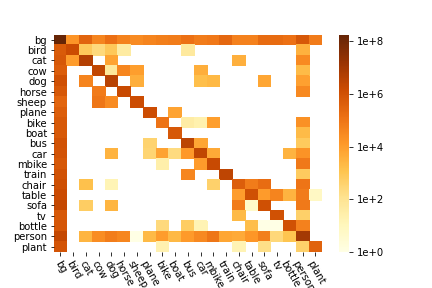}
        \caption{With prediction filtering}
        \label{fig:confusion-filtered}
    \end{subfigure}
    \vspace*{-0.8mm}
    \caption{Pixel-level confusion matrices for DeeplabV1 models trained on $1{,}464$ pixel-level labels. Each entry shows the number of pixels (on a logarithmic scale; $0$ values are plotted as if they were $1$) whose true label is given according to the row, and whose predicted label is that in the column. Labels are sorted into rough categories to show block structure.}
     \vspace*{-1.0mm}
    \label{fig:confusion-matrices}
\end{figure*}

\begin{figure*}[h!]
    \centering
    \includegraphics[width=0.92\textwidth]{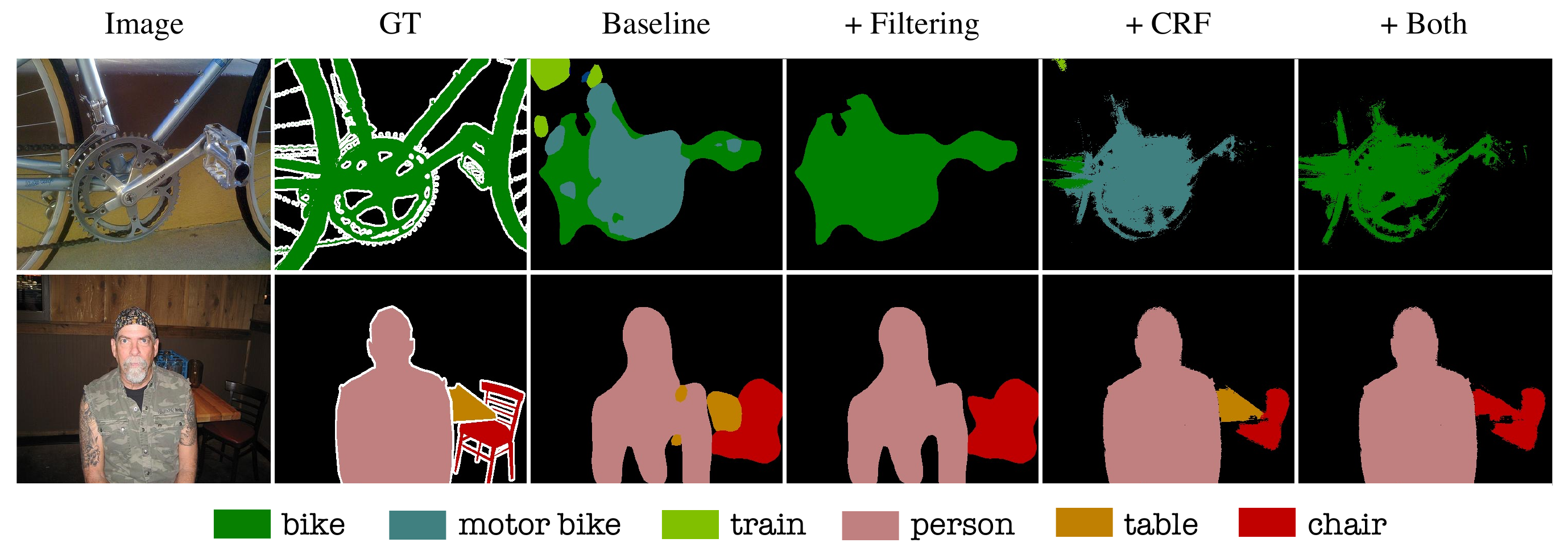}
    \vspace*{-1.5mm}
    \caption{Qualitative results, the top row of a successful case and bottom a failure
    using DeeplabV1 trained on $1{,}464$ pixel-labeled images.
    }
     \vspace*{-3.0mm}
    \label{fig:qualitative}
\end{figure*}

\paragraph{Comparison with state-of-the-art.}
In \cref{tbl:sota}, we compare the performance of prediction filtering to existing methods when $\lvert\Dpixel\rvert$ is $1{,}464$.
We reproduce results for Deeplab, OCRNet, WSSL, CCT, DualNet, and AdvCAM.\footnote{%
Our AdvCAM result is substantially worse than reported by \citet{AdvCAM:2021Lee}, because we did not apply tricks such as multi-scale and flipping in inference time for fair comparison with the other state-of-the-art methods.}
Among VGG-based methods, DeeplabV1 with prediction filtering outperforms the other methods without using any additional information.
Similarly, filtering also improves models with stronger backbones, though the margin of improvement is less dramatic since the baseline is better.
Prediction filtering can help even when it does not involve adding any new training data:
it significantly helps WSSL, CCT, AdvCAM, and DualNet although they already use weakly-labeled data.
It is worth highlighting that simply adding prediction filtering to the DeeplabV3-W baseline achieves the new highest performance, slightly higher than DualNet (with a ResNet-W backbone) with prediction filtering, both of which are notably better than the previous state-of-the-art (DualNet with a ResNet-W backbone without prediction filtering).
Prediction filtering on top of DualNet also sets the new state-of-the-art for VGG-based models.





\paragraph{Results on the test set.}
We further evaluate prediction filtering on the test set of VOC2012.
In \cref{tbl:test}, we provide the performance of DeeplabV1 and DeeplabV3-W on the test set, as well as with prediction filtering applied.
We can observe that prediction filtering improves the intersection-over-union (IoU) scores for most of the $21$ classes,
leading to significant improvements in terms of mIoU (as on the validation set).

\paragraph{Various levels of supervision.}
\Cref{fig:various_supervision} shows the segmentation performance of DeeplabV1, DualNet with a VGG backbone, and DeeplabV3-W trained on $200$, $500$, $800$ and $1{,}464$ images with pixel labels, with and without prediction filtering.
The blue (bottom) line shows performance of the base model;
orange (middle) shows with prediction filtering;
green (top) is with oracle filtering, which upper-bounds the possibility of improvement of prediction filtering with a better classifier.
For smaller numbers of pixel labels, the performance gain from prediction filtering is drastically larger.
For example, at $800$ images with pixel labels,
DeeplabV1 goes from $56.8 \rightarrow 64.4$,
DeeplabV3-W from $72.3 \rightarrow 74.3$.
The improvements for $200$ pixel labels are  $12.4$, $1.8$, and $4.8$.

\paragraph{Relationship with CRF.}
CRFs adjust the prediction for each pixel by encouraging nearby, similar-colored pixels to have the same label. This usually improves the segmentation performance by refining detailed boundaries and making class predictions across an object more consistent.
The latter role overlaps to some extent with prediction filtering.
If the size of the wrong prediction is large, though, a CRF might expand the area of the wrong prediction, rather than remove it.
\Cref{tbl:crf_and_partial}, as well as the qualitative results to come shortly, shows that the methods complement one another.

\paragraph{Without image-level labels.}
Although image-level labels are easier to obtain than pixel-level labels, annotation effort is still nontrivial.
Our hypothesis about why prediction filtering works, however, is largely that classifiers ``looks at images differently'' than segmentation networks do.
It thus might help even if it does not use any additional data: $\Dpixel$ and $\Dimage$ contain the same images.
\cref{tbl:crf_and_partial} (right) shows this is the case.
Even without introducing any actual new information, prediction filtering still improves mIoU significantly.




\paragraph{Changes between classes.}
In \cref{sec:problem}, we showed some qualitative evidence that the segmentation network with low pixel labels tends to be confused between similar classes,
and that prediction filtering can help to compensate it by looking at other parts of the image.
To further demonstrate this, \cref{fig:confusion-matrices} shows the pixel-level confusion matrix for a DeeplabV1 model with CRF before and after prediction filtering.
\cref{fig:confusion-matrices}(a) shows a strong block structure where pixels from one animal class are often confused for another animal class, or vehicles for vehicles.
In \cref{fig:confusion-matrices}(b), prediction filtering has dramatically reduced these types of mistakes.


\paragraph{Qualitative results.}
\Cref{fig:qualitative} shows a success (top) and failure (bottom) for prediction filtering.
At top, an object is mostly incorrectly predicted -- \labelname{bike} as \labelname{motor bike} -- and CRF only makes this worse.
The classifier in filtering, however, correctly identifies there is no \labelname{motor bike}, and the model's ``second choice'' prediction is largely correct.
In the failure case, an object (\labelname{table}) is mostly occluded;
the segmentation model still identifies it, but the classifier misses it.

\paragraph{Additional ablation studies.}
The appendix provides further experiments, including alternative approaches to prediction filtering, performance with model variants, and a runtime comparison between prediction filtering and CRFs. 


\section{Conclusion}
\label{sec:conclusion}
Most existing semi-weakly supervised semantic segmentation algorithms exploit the pseudo-labels extracted from a classifier.
Doing so, however, requires a complicated architecture and extensive hyperparameter tuning on fully-supervised validation sets.
We propose \textit{prediction filtering}, a simple post-processing method that only considers the classes for segmentation a classifier is confident are present.
Our experiments demonstrated adding this method to baselines achieves the new highest performance on PASCAL VOC in SWSSS regimes, and adding it to existing SWSSS algorithms uniformly improves their performance.
We expect prediction filtering can become a standard post-processing method for segmentation, along with CRFs, at least when a relatively large number of weakly-labeled images are available and the portion of class labels present in most images is low.

\bibliographystyle{named}
\bibliography{ijcai22}

\begin{thebibliography}{}

\bibitem[\protect\citeauthoryear{Ahn and Kwak}{2018}]{PSA:2018Ahn}
J.~Ahn and S.~Kwak.
\newblock {Learning pixel-level Semantic Affinity with Image-level Supervision
  for Weakly Supervised Semantic Seg.}
\newblock In {\em CVPR}, 2018.

\bibitem[\protect\citeauthoryear{Chen \bgroup \em et al.\egroup
  }{2015}]{DeeplabV1:2015Chen}
L.-C. Chen, G.~Papandreou, I.~Kokkinos, K.~Murphy, and A.~L. Yuille.
\newblock {Semantic Image Segmentation with Deep Convolutional Nets and Fully
  Connected {CRF}s}.
\newblock In {\em ICLR}, 2015.

\bibitem[\protect\citeauthoryear{Chen \bgroup \em et al.\egroup
  }{2017}]{DeeplabV3:2017Chen}
L.-C. Chen, G.~Papandreou, F.~Schroff, and H.~Adam.
\newblock {Rethinking Atrous Convolution for Semantic Image Segmentation}.
\newblock In {\em CVPR}, 2017.

\bibitem[\protect\citeauthoryear{Chen \bgroup \em et al.\egroup
  }{2018}]{DeeplabV3+:2018Chen}
L.-C. Chen, Y.~Zhu, G.~Papandreou, F.~Schroff, and H.~Adam.
\newblock {Encoder-decoder with Atrous Separable Conv. for Semantic Image Seg.}
\newblock In {\em ECCV}, 2018.

\bibitem[\protect\citeauthoryear{Chen \bgroup \em et al.\egroup
  }{2020}]{Boundary:2020Chen}
L.~Chen, W.~Wu, C.~Fu, X.~Han, and Y.~Zhang.
\newblock {Weakly Supervised Semantic Segmentation with Boundary Exploration}.
\newblock In {\em ECCV}, 2020.

\bibitem[\protect\citeauthoryear{Choe and Shim}{2019}]{ADL:Choe2019}
J.~Choe and H.~Shim.
\newblock {Attention-Based Dropout Layer for Weakly Supervised Object
  Localization}.
\newblock In {\em CVPR}, 2019.

\bibitem[\protect\citeauthoryear{Choe \bgroup \em et al.\egroup
  }{2020}]{EvalWSOL:Choe2020}
J.~Choe, S.~Oh, S.~Lee, S.~Chun, Z.~Akata, and H.~Shim.
\newblock {Evaluating Weakly Supervised Object Localization Methods Right}.
\newblock In {\em CVPR}, 2020.

\bibitem[\protect\citeauthoryear{Dang \bgroup \em et al.\egroup
  }{2022}]{vessel2022dang}
V.~Dang, F.~Galati, R.~Cortese, G.~Di~Giacomo, V.~Marconetto, P.~Mathur,
  K.~Lekadir, M.~Lorenzi, F.~Prados, and M.~Zuluaga.
\newblock {Vessel-CAPTCHA: An Efficient Learning Framework for Vessel
  Annotation and Segmentation}.
\newblock {\em Medical Image Analysis}, 2022.

\bibitem[\protect\citeauthoryear{Everingham \bgroup \em et al.\egroup
  }{2015}]{VOC:Everingham2015}
M.~Everingham, A.~Eslami, L.~Gool, C.~Williams, J.~Winn, and A.~Zisserman.
\newblock {{P}ascal Vis. Obj. Class Challenge: A Retrospective}.
\newblock {\em IJCV}, 2015.

\bibitem[\protect\citeauthoryear{He \bgroup \em et al.\egroup
  }{2016}]{Resnet:He2015}
K~He, X.~Zhang, S.~Ren, and J.~Sun.
\newblock {Deep Residual Learning for Image Recognition}.
\newblock In {\em CVPR}, 2016.

\bibitem[\protect\citeauthoryear{Huang \bgroup \em et al.\egroup
  }{2018}]{DSRG:2018Huang}
Z.~Huang, X.~Wang, J.~Wang, W.~Liu, and J.~Wang.
\newblock {Weakly-supervised Semantic Seg. Network with Deep Seeded Region
  Growing}.
\newblock In {\em CVPR}, 2018.

\bibitem[\protect\citeauthoryear{Jiang \bgroup \em et al.\egroup
  }{2019}]{OAA:2019Jiang}
P.-T. Jiang, Q.~Hou, Y.~Cao, M.-M. Cheng, Y.~Wei, and H.-K. Xiong.
\newblock {Integral Object Mining via Online Attention Accumulation}.
\newblock In {\em ICCV}, 2019.

\bibitem[\protect\citeauthoryear{Kr{\"a}henb{\"u}hl and
  Koltun}{2011}]{CRF:Krahenbuhl2011}
P.~Kr{\"a}henb{\"u}hl and V.~Koltun.
\newblock {Efficient Inference in Fully Connected {CRF}s with {G}aussian Edge
  Potentials}.
\newblock In {\em NIPS}, 2011.

\bibitem[\protect\citeauthoryear{Lai \bgroup \em et al.\egroup
  }{2021}]{semi_context_const2021lai}
X.~Lai, Z.~Tian, L.~Jiang, S.~Liu, H.~Zhao, L.~Wang, and J.~Jia.
\newblock {Semi-Supervised Semantic Seg. with Directional Context-aware
  Consistency}.
\newblock In {\em CVPR}, 2021.

\bibitem[\protect\citeauthoryear{Lee \bgroup \em et al.\egroup
  }{2019}]{FickleNet:2019Lee}
J.~Lee, E.~Kim, S.~Lee, J.~Lee, and S.~Yoon.
\newblock {Ficklenet: Weakly and Semi-supervised Semantic Image Segmentation
  Using Stochastic Inference}.
\newblock In {\em CVPR}, 2019.

\bibitem[\protect\citeauthoryear{Lee \bgroup \em et al.\egroup
  }{2021}]{AdvCAM:2021Lee}
J.~Lee, E.~Kim, and S.~Yoon.
\newblock {Anti-Adversarially Manipulated Attributions for Weakly and
  Semi-Supervised Semantic Segmentation}.
\newblock In {\em CVPR}, 2021.

\bibitem[\protect\citeauthoryear{Li \bgroup \em et al.\egroup
  }{2018}]{GAIN:2018Li}
K.~Li, Z.~Wu, K.~Peng, J.~Ernst, and Y.~Fu.
\newblock {Tell Me Where to Look: Guided Attention Inference Network}.
\newblock In {\em CVPR}, 2018.

\bibitem[\protect\citeauthoryear{Lin \bgroup \em et al.\egroup
  }{2014}]{COCO:Lin2014}
T.~Lin, M.~Maire, S.~Belongie, J.~Hays, P.~Perona, D.~Ramanan, P.~Doll{\'a}r,
  and C.~Zitnick.
\newblock {MS COCO: Common Objects in Context}.
\newblock In {\em ECCV}, 2014.

\bibitem[\protect\citeauthoryear{Luo and Yang}{2020}]{Dual:2020Luo}
W.~Luo and M.~Yang.
\newblock {Semi-supervised Semantic Segmentation via Strong-weak Dual-branch
  Network}.
\newblock In {\em ECCV}, 2020.

\bibitem[\protect\citeauthoryear{Ouali \bgroup \em et al.\egroup
  }{2020}]{CCT:2020Ouali}
Y.~Ouali, C.~Hudelot, and M.~Tami.
\newblock {Semi-supervised Semantic Segmentation with Cross-consistency
  Training}.
\newblock In {\em CVPR}, 2020.

\bibitem[\protect\citeauthoryear{Papandreou \bgroup \em et al.\egroup
  }{2015}]{WSSL:2015Papandreou}
G.~Papandreou, L.~Chen, K.~P Murphy, and A.~L Yuille.
\newblock {Weakly– and Semi-Supervised Learning of a Deep Convolutional
  Network for Semantic Image Segmentation}.
\newblock In {\em ICCV}, 2015.

\bibitem[\protect\citeauthoryear{Selvaraju \bgroup \em et al.\egroup
  }{2017}]{Gradcam:Selvaraju2016}
R.~R. Selvaraju, M.~Cogswell, A.~Das, R.~Vedantam, D.~Parikh, and D.~Batra.
\newblock {{Grad-Cam}: Visual Explanations From Deep Networks via
  Gradient-Based Localization}.
\newblock In {\em ICCV}, 2017.

\bibitem[\protect\citeauthoryear{Simonyan and
  Zisserman}{2015}]{VGG:Simonyan2014}
K.~Simonyan and A.~Zisserman.
\newblock {Very Deep Convolutional Networks for Large-Scale Image Recognition}.
\newblock In {\em ICLR}, 2015.

\bibitem[\protect\citeauthoryear{Singh and Lee}{2017}]{HaS:Singh2017}
K.~Singh and Y.~Lee.
\newblock {Hide-And-Seek: Forcing a network to be meticulous for
  weakly-supervised object and action localization}.
\newblock In {\em ICCV}, 2017.

\bibitem[\protect\citeauthoryear{Souly \bgroup \em et al.\egroup
  }{2017}]{SemiGAN2017Souly}
N.~Souly, C.~Spampinato, and M.~Shah.
\newblock {Semi and Weakly Supervised Semantic Segmentation Using Generative
  Adversarial Network}.
\newblock {\em ICCV}, 2017.

\bibitem[\protect\citeauthoryear{Strudel \bgroup \em et al.\egroup
  }{2021}]{OCRNet:2021Strudel}
R.~Strudel, R.~Garcia, I.~Laptev, and C.~Schmid.
\newblock {Segmenter: Transformer for Semantic Segmentation}.
\newblock In {\em ICCV}, 2021.

\bibitem[\protect\citeauthoryear{Sun \bgroup \em et al.\egroup
  }{2020}]{MCIS:2020Sun}
G.~Sun, W.~Wang, J.~Dai, and L.~Van~Gool.
\newblock {Mining Cross-image Semantics for Weakly Supervised Semantic
  Segmentation}.
\newblock In {\em ECCV}, 2020.

\bibitem[\protect\citeauthoryear{Wang \bgroup \em et al.\egroup
  }{2020}]{SEAM:2020Wang}
Y.~Wang, J.~Zhang, M.~Kan, S.~Shan, and X.~Chen.
\newblock {Self-supervised Equivariant Attention Mechanism for Weakly
  Supervised Sem. Seg.}
\newblock In {\em CVPR}, 2020.

\bibitem[\protect\citeauthoryear{Wang \bgroup \em et al.\egroup
  }{2021}]{HRNet2020Wang}
J.~Wang, K.~Sun, T.~Cheng, B.~Jiang, C.~Deng, Y.~Zhao, D.~Liu, Y.~Mu, M.~Tan,
  X.~Wang, W.~Liu, and B.~Xiao.
\newblock {Deep High-Resolution Representation Learning for Visual
  Recognition}.
\newblock {\em IEEE T-PAMI}, 2021.

\bibitem[\protect\citeauthoryear{Wei \bgroup \em et al.\egroup
  }{2018}]{MDC:2018Wei}
Y.~Wei, H.~Xiao, H.~Shi, Z.~Jie, J.~Feng, and T.~S Huang.
\newblock {Revisiting Dilated Convolution: A Simple Approach for Weakly- and
  Semi-supervised Semantic Segmentation}.
\newblock In {\em CVPR}, 2018.

\bibitem[\protect\citeauthoryear{Wu \bgroup \em et al.\egroup
  }{2019}]{Wider:2019Wu}
Z.~Wu, C.~Shen, and A.~Hengel.
\newblock {Wider or Deeper: Revisiting the Resnet Model for Visual
  Recognition}.
\newblock {\em Pattern Recognition}, 2019.

\bibitem[\protect\citeauthoryear{Yu and Koltun}{2016}]{Dilated:2015Yu}
F.~Yu and V.~Koltun.
\newblock {Multi-scale Context Aggregation by Dilated Convolutions}.
\newblock In {\em ICLR}, 2016.

\bibitem[\protect\citeauthoryear{Zhang \bgroup \em et al.\egroup
  }{2018}]{ACoL:Zhang2018}
X.~Zhang, Y.~Wei, J.~Feng, and T.~S. Yang, Y.~Huang.
\newblock {Adversarial Complementary Learning for Weakly Supervised Object
  Localization}.
\newblock In {\em CVPR}, 2018.

\bibitem[\protect\citeauthoryear{Zhou \bgroup \em et al.\egroup
  }{2016}]{CAM:Zhou2015}
B.~Zhou, A.~Khosla, A.~Lapedriza, A.~Oliva, and A.~Torralba.
\newblock {Learning Deep Features for Discriminative Localization}.
\newblock In {\em CVPR}, 2016.

\end{thebibliography}

\clearpage
\appendix
\section*{Appendices}

In the remainder of the paper, we provide additional experimental results, as well as detailed descriptions for the experimental setup.
\begin{itemize}
    \item \cref{sec:pseudo_code} provides pseudocode for training and inference of models with prediction filtering.
    \item \cref{sec:comparison_training} compares the time complexity of training time between prediction filtering and recent SWSSS algorithms.
    \item \cref{sec:comparison_hyperparameters} compares prediction filtering with the state-of-the-art models in terms of mIoU and the number of additional hyperparameters used.
    \item \cref{sec:ensemble} provides the performance of ensemble of two segmentation networks.
    \item \cref{sec:runtime} compares the runtime of prediction filtering with CRF.
    \item \cref{sec:different_cls} provides results when using a (worse) VGG16 classification network for prediction filtering, rather than the ResNet50 classifier of the main paper.
    \item \cref{sec:global_context} analyzes the interaction of prediction filtering with a segmentation network's ability to capture the global context.
    \item \cref{sec:alternatives} provides alternative approaches to prediction filtering, and compares them based on additional experiment results.
    \item \cref{sec:subgroup} compares the performance with and without prediction filtering on sub-groups in terms of the number of classes and the size of classes in an image.
    \item \cref{sec:sensitivity} shows that the segmentation performance is not very sensitive to filtering threshold.
    \item \cref{sec:reproducibility} provides detailed numbers for hyperparameters used in the experiments, and a link to (anonymized) source code for prediction filtering.
    \item \cref{sec:additional_qualitative} shows additional examples of both successes and failures of the model.
\end{itemize}

\begin{figure*}[th!]
    \centering
    \includegraphics[width=0.95\textwidth]{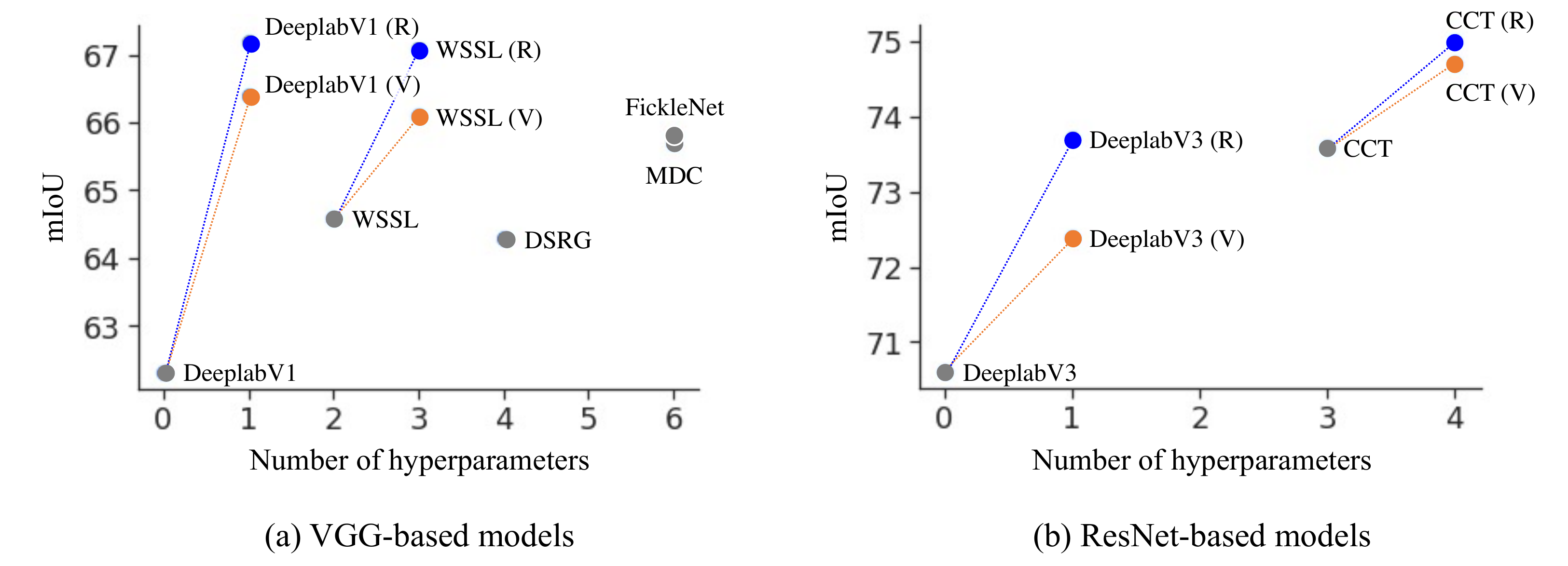}
    \caption{Performance of the baselines and state-of-the-art models with respect to the number of hyperparameters. (V) in orange represents prediction filtering with VGG16, while (R) in blue represents the prediction filtering using ResNet50.}
    \label{fig:hyperparameters}
    \vspace*{3mm}
\end{figure*}

\section{Pseudocode}
\label{sec:pseudo_code}

Here, we provide pseudocode describing training and inference with prediction filtering.  
\begin{algorithm}
\caption{Prediction filtering}\label{alg:pred_filter}
\KwInput{\\
\Indp $D_{pixel}$ -- training images with pixel-level labels \\
      $D_{image}$ -- training images with image-level labels \\
      $D_{test}$ -- test images \\
      $\tau$ -- a threshold for prediction filtering\\
\Indm
}
\KwRequire{\\
\Indp $f$ -- a segmentation network \\
      $g$ -- a classification network \\
\Indm
}
\KwOutput{\\
\Indp Prediction results on $D_{test}$ \\
\Indm
}

Train $f$ on $D_{pixel}$\;
Train $g$ on $D_{image}$\;
\For{\text{each image} $x$ \textbf{\textup{in}} $D_{test}$} {
  Filter the set of candidate classes, $\mathcal{K} = \{ c: g(x)_c > \tau \}$ \;
  \For{\text{each spatial location } (h, w) }{
    Predict $\hat{c}_{h,w}(x) = \argmax_{c \in \mathcal{K}} f(x)_{(c,h,w)}$;
  }
}
\end{algorithm}

\section{Comparison of Time Complexity in Training time}
\label{sec:comparison_training}
At training time, prediction filtering requires training an additional classification network.
The overhead here, however, is substantially smaller than most SWSSS algorithms, most of which require the following steps:
\begin{enumerate}
    \item Train a classification network to generate “seed” regions of objects.
    \item Train a region growing network, such as DSRG [10], PSA [9], or IRN [A], to generate class activation maps (CAMs).
    \item Apply a post-processing method like CRF [20] to generate the final pseudo-labels.
    \item Train a segmentation network jointly on pixel-level labels and the generated pseudo-labels.
\end{enumerate}

As each step depends on its previous step, none of the above steps are parallelizable.
Also, the quality of the final pseudo-labels -- which cannot be measured until step 4 -- is highly sensitive to hyperparameters (\cref{fig:problems}(c)) chosen in steps 2 and 3.
Thus hyperparameter tuning requires repeating all of steps 2, 3, and 4 many times.

By comparison, with prediction filtering we can train the segmentation and classification network in parallel.
When the number of images with image-level labels is much larger than that with pixel-level labels (as in the usual SWSSS setting), the total training wall time is no more than the time it would take to train a classification network (the very first step of most SWSSS algorithms).
We do not provide the exact training time of our method compared to the previous methods since it is difficult to fairly compare hyperparameter tuning cost.
Instead, we provide the number of hyperparameters required for different algorithms in \cref{sec:comparison_hyperparameters}.

\section{Comparison of Hyperparameters}
\label{sec:comparison_hyperparameters}

As mentioned in the main paper, the pseudo-labels used by most SWSSS algorithms require several additional hyperparameters.
In \cref{fig:hyperparameters}, we compare the segmentation performance of different models (in mIoU) with respect to the number of additional hyperparameters used to provide extra supervision from image-level labels, which require tuning through full supervision in a validation set.
For clarity, we also list the hyperparameters considered for every model.
Note that we do not consider hyperparameters used to train a classifier, since choosing those does not require any pixel-level labels.
\begin{itemize}
    \item Prediction filtering (\textbf{(R)} and \textbf{(V)} in \cref{fig:hyperparameters}, for ResNet50 and VGG16 filter classifiers, respectively) require one additional threshold to determine the presence of a class in an image.
    \item \textbf{WSSL}~\cite{WSSL:2015Papandreou} requires two additional hyperparameters; two coefficients to generate pseudo-labels based on image-level labels.
    \item \textbf{MDC}~\cite{MDC:2018Wei} requires six additional hyperparameters. First, it requires the number of dilation blocks. For each dilation block, dilation rate also needs to be determined; we count them as three since three blocks are used in the original paper). Lastly, two thresholds are required to determine foreground and background regions from class activation maps (CAMs) and background cues.
    \item \textbf{DSRG}~\cite{DSRG:2018Huang} requires four additional hyperparameters; two thresholds to determine foreground and background regions and two thresholds to determine similarity of different regions (super-pixels).
    \item \textbf{FickleNet}~\cite{FickleNet:2019Lee} requires six additional hyperparameters. As FickleNet is based on DSRG, it requires four hyperparameters that are used in DSRG. In addition, the number of localization maps and dropout rate are required to generate seed regions for DSRG.
    \item \textbf{CCT}~\cite{CCT:2020Ouali} requires three additional hyperparameters. As with many others, it requires two thresholds to determine foreground and background regions. Also, to use a CRF to generate pseudo-labels, it needs to convert predictions into probability which is determined by one hyperparameter.
\end{itemize}

As shown in \cref{fig:hyperparameters}, DeeplabV1 and WSSL with prediction filtering (V) is comparable to other state-of-the-art SWSSS models with only one additional hyperparameter.
More importantly, with prediction filtering (R), DeeplabV1 and WSSL outperform the other VGG-based models by a significant margin.
Similarly, DeeplabV3 with prediction filtering is comparable with CCT with two fewer hyperparameters.
By adding prediction filtering, the performance of CCT is further boosted up with one additional hyperparameter.

\section{Comparison with Ensemble Model}
\label{sec:ensemble}
The proposed prediction filtering can be considered as a particular form of an ensemble between two networks (segmentation and classification networks).
Another ensemble that could be useful to compare against is two segmentation models (and no classifier), in the case where there are no image-only labels, as this might help to show what might be due to ensemble effect.
Thus, we train two DeeplabV1 segmentation networks on 1,464 training images.
We initialized them with ImageNet-pretrained weights for the earlier layers and randomly initialized weights for the last few layers. (Using entirely random weights gives poor performance, since there is not enough training data.) At inference time, we averaged the output probabilities, giving slightly better performances than the baselines (without CRF -- baseline: $57.2$ \vs ensemble: $57.7$, with CRF -- baseline: $61.8$ \vs ensemble: $62.5$) although it adds much memory and time complexity as a classification network.
We conjecture this is because, despite random initialization, two segmentation networks look at the data from a similar perspective, unlike a classification network.

\section{Comparison of Runtime between Prediction Filtering and CRF}
\label{sec:runtime}
In Table~\ref{tbl:crf_and_partial} (Left), we demonstrate prediction filtering (without a CRF) is significantly better than CRF on both $500$ and $1{,}464$ images with pixel-level labels regardless of backbone structures.
In terms of runtime, using CRFs can be computationally quite expensive – here, prediction filtering can be quite handy.
Using a NVIDIA P100 GPU, inference time on the full validation set with prediction filtering (no CRF) is $1$ min $58$ sec ($0.08$ sec / image), while with CRF (no filtering) it is $50$ min $48$ sec ($2.1$sec / image); for context, baseline inference with no post-processing takes about $1$ min $16$ sec ($0.05$ sec / image). 
Hence, prediction filtering provides about as much accuracy as CRF with far less computational overhead.

\begin{table*}[tp]
\centering
\fontsize{9.0}{12.0}\selectfont
\begin{tabular}{c|c|l|c|c|c|l}
\hline
Backbone & Add. 9.1K Images & \multicolumn{1}{c|}{Method} & Bkg. Cues & CRF & Pred. Filter & $\!$mIoU \\
\hline\hline
\multirow{16}{*}{VGG} 
  & \multirow{3}{*}{--} & DeeplabV1~\cite{DeeplabV1:2015Chen}  & -- & \cm & -- &  61.8 \\
  & & DeeplabV1~\cite{DeeplabV1:2015Chen}  & -- & \cm & VGG & 63.4 \\
  & & DeeplabV1~\cite{DeeplabV1:2015Chen}  & -- & \cm & ResNet & 64.9 \\ \hhline{*{1}{|~}*{6}{|-}}
  & \multirow{10}{*}{Image-level} & \cellcolor{myGray}DeeplabV1~\cite{DeeplabV1:2015Chen}  & \cellcolor{myGray}-- & \cellcolor{myGray}\cm & \cellcolor{myGray}VGG & \cellcolor{myGray}66.1 \\
  & \multirow{8}{*}{Image-level} & \cellcolor{myGray}DeeplabV1~\cite{DeeplabV1:2015Chen}  & \cellcolor{myGray}-- & \cellcolor{myGray}\cm & \cellcolor{myGray}ResNet & \cellcolor{myGray}67.4 \\
  & & WSSL~\cite{WSSL:2015Papandreou}  & --  & \cm & -- & 64.6 \\
  & & \cellcolor{myGray}WSSL~\cite{WSSL:2015Papandreou} & \cellcolor{myGray}-- & \cellcolor{myGray}\cm & \cellcolor{myGray}VGG & \cellcolor{myGray}66.1 \\
  & & \cellcolor{myGray}WSSL~\cite{WSSL:2015Papandreou} & \cellcolor{myGray}-- & \cellcolor{myGray}\cm & \cellcolor{myGray}ResNet & \cellcolor{myGray}67.1 \\
  & & \citet{SemiGAN2017Souly}  & -- & -- & -- & 65.8$^*$ \\
  & & GAIN~\cite{GAIN:2018Li}  & \cm & \cm & -- &  60.5$^*$ \\
  & & MDC~\cite{MDC:2018Wei}  & \cm   & \cm & -- & 65.7$^*$  \\
  & & DSRG~\cite{DSRG:2018Huang}  & \cm  &  \cm & -- & 64.3$^*$ \\
  & & FickleNet~\cite{FickleNet:2019Lee} & \cm  & \cm & -- & 65.8$^*$ \\ \hhline{*{1}{|~}*{6}{|-}}
  & \multirow{3}{*}{Pixel-level} & DeeplabV1~\cite{DeeplabV1:2015Chen}  & -- & \cm & -- & 69.0  \\
  &  & DeeplabV1~\cite{DeeplabV1:2015Chen} & -- & \cm & VGG & 69.7 \\
  &  & DeeplabV1~\cite{DeeplabV1:2015Chen} & -- & \cm & ResNet & 71.6  \\ \Xhline{2\arrayrulewidth}
\multirow{12}{*}{ResNet}
  & \multirow{4}{*}{--} & DeeplabV3~\cite{DeeplabV3:2017Chen} & -- & \cm & -- & 72.4 \\
  & & DeeplabV3~\cite{DeeplabV3:2017Chen} & -- & \cm & VGG & 73.1 \\
  & & DeeplabV3~\cite{DeeplabV3:2017Chen} & -- & \cm & ResNet & 73.7 \\
  & & \citet{semi_context_const2021lai} & -- & -- & -- & 74.5 \\
  \hhline{*{1}{|~}*{6}{|-}}
  & \multirow{5}{*}{Image-level} & \cellcolor{myGray}DeeplabV3~\cite{DeeplabV3:2017Chen} & \cellcolor{myGray}-- & \cellcolor{myGray}\cm & \cellcolor{myGray}VGG & \cellcolor{myGray}73.8 \\
  & & \cellcolor{myGray}DeeplabV3~\cite{DeeplabV3:2017Chen} & \cellcolor{myGray}-- & \cellcolor{myGray}\cm & \cellcolor{myGray}ResNet & \cellcolor{myGray}75.3 \\ 
  & & CCT~\cite{CCT:2020Ouali} & -- & \cm & -- & 74.7 \\
  & & \cellcolor{myGray}CCT~\cite{CCT:2020Ouali} & \cellcolor{myGray}-- & \cellcolor{myGray}\cm & \cellcolor{myGray}VGG & \cellcolor{myGray}75.9 \\
  & & \cellcolor{myGray}CCT~\cite{CCT:2020Ouali} & \cellcolor{myGray}-- & \cellcolor{myGray}\cm & \cellcolor{myGray}ResNet & \cellcolor{myGray}76.0 \\
\hhline{*{1}{|~}*{6}{|-}}
  & \multirow{3}{*}{Pixel-level} & DeeplabV3~\cite{DeeplabV3:2017Chen} & -- & \cm &-- & 77.4 \\ 
  & & DeeplabV3~\cite{DeeplabV3:2017Chen} & -- & \cm & VGG & 77.8 \\
  & & DeeplabV3~\cite{DeeplabV3:2017Chen} & -- & \cm & ResNet & 77.8 \\ \Xhline{2\arrayrulewidth}
\end{tabular}
\caption{Comparison of the state-of-the-art methods on $1{,}464$ images with pixel labels (the original VOC2012 training set). The ``Add. 9.1K Images'' column gives which type of supervision is used for the 9.1K additional images (augmented dataset). ``Pred. Filter'' column gives which classifier is used for prediction filtering.}
\label{tbl:sota_with_vgg}
\end{table*}

\newcolumntype{C}{>{\centering\arraybackslash}p{2.6em}}
\begin{table*}[tp]
\centering
\fontsize{9.0}{12.0}\selectfont
\begin{tabu}{c|c|CCCC|CCCC}
\hline
\multicolumn{1}{c|}{\multirow[c]{2}{*}{Method}} & \multirow[c]{2}{*}{Pred. Filter} & \multicolumn{4}{c|}{Without CRF ($M$)} & \multicolumn{4}{c}{With CRF ($M$)} \\
\cline{3-10}
 & & 200 & 500 & 800 & 1,464 & 200 & 500 & 800 & 1,464  \\
\hline
\hline
\multirow{4}{*}{DeeplabV1~\cite{DeeplabV1:2015Chen}} 
& --    & 43.3 & 49.9 & 52.7 & 57.2 & 46.3 & 53.9 & 56.8 & 61.8  \\
& VGG   & 55.7 & 59.1 & 61.1 & 63.2 & 57.4 & 61.4 & 63.4 & 66.1 \\
& ResNet& 57.1 & 61.4 & 62.2 & 64.6 & 58.8 & 63.6 & 64.4 & 67.4 \\
& Oracle& 59.9 & 64.6 & 65.6 & 68.1 & 61.3 & 66.5 & 67.5 & 70.6  \\
\hline
\multirow{4}{*}{DeeplabV3~\cite{DeeplabV3:2017Chen}}
& --    & 38.2  & 57.0 & 62.6 & 70.6 & 39.3 & 58.4 & 65.1 & 72.4 \\
& VGG   & 50.7 & 63.3 & 66.7 & 72.4 & 51.0 & 65.0 & 68.3 & 73.8 \\
& ResNet& 52.1 & 64.8 & 67.6 & 74.0 & 52.3 & 65.9 & 69.1 & 75.3 \\
& Oracle& 53.9 & 67.8 & 70.3 & 76.8 & 53.9 & 68.7 & 71.7 & 77.7 \\
\hline
\multirow{4}{*}{CCT~\cite{CCT:2020Ouali}}
& --    &59.9 & 66.4 & 68.7 & 73.4 & 61.2 & 67.7 & 69.9 & 74.7 \\
& VGG   & 63.4 & 68.0 & 69.9 & 74.7 & 64.4 & 68.8 & 71.0 & 75.9 \\
& ResNet& 64.2 & 68.9 & 70.6 & 74.8 & 65.2 & 69.9 & 71.7 & 76.0 \\
& Oracle& 65.9 & 71.1 & 72.6 & 77.4 & 66.8 & 72.0 & 73.5 & 78.4 \\
\Xhline{2\arrayrulewidth}
\end{tabu}
\caption{Performance of the prediction filtering on various models and levels of supervision ($M$).}
\label{tbl:additional_various_supervision}
\end{table*}

\begin{figure*}[h!]
    \centering
    \includegraphics[width=\textwidth]{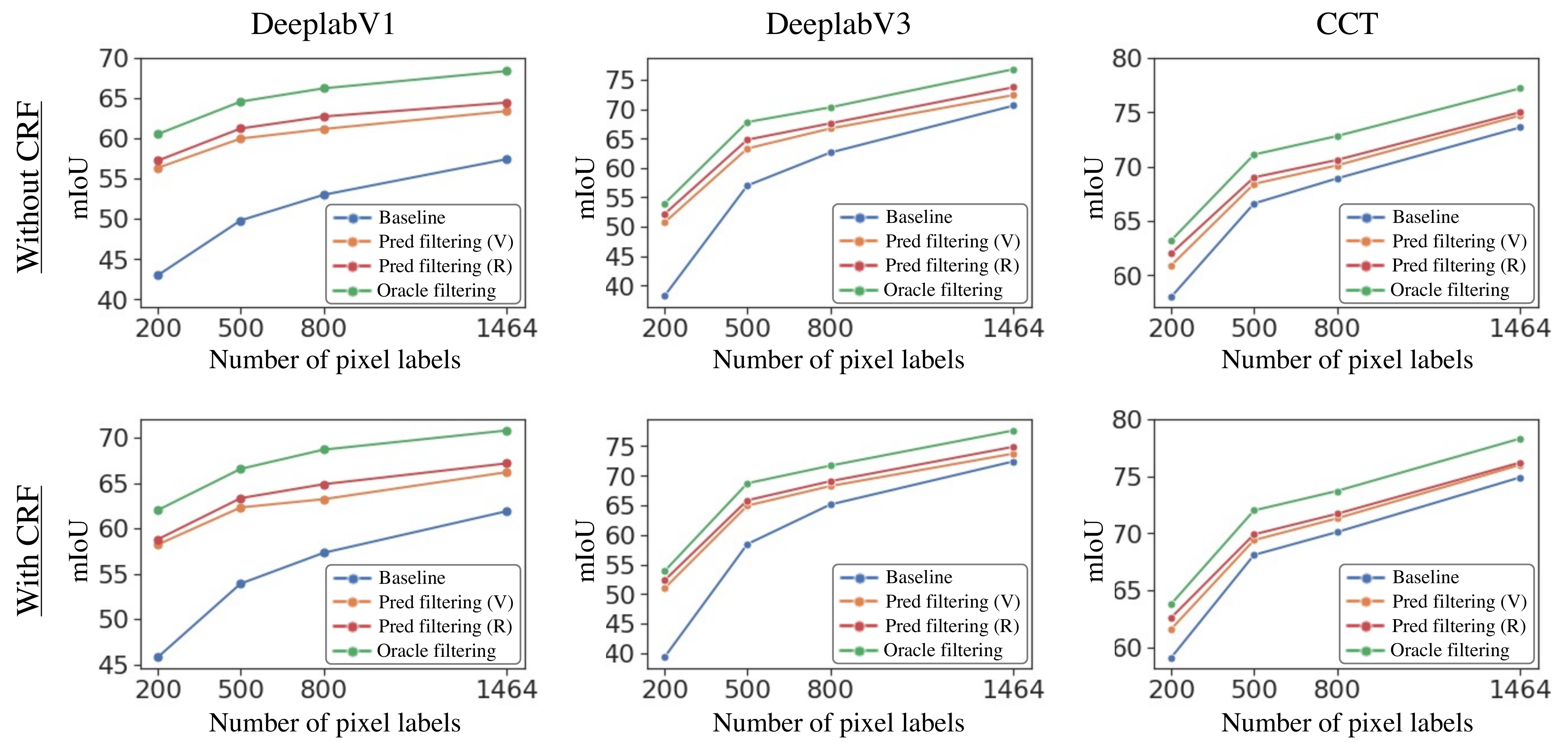}
    \caption{Performance of the prediction filtering on various models and levels of supervision.}
    \label{fig:additional_various_supervision}
    \vspace*{4mm}
\end{figure*}

\section{Different Classification Network for Prediction Filtering}
\label{sec:different_cls}
All experiments in the main paper used a ResNet50~\cite{Resnet:He2015} classifier for prediction filtering, due to its strong performance.
But, in some environments with limited memory and space, a lighter classifier may be useful.
Here, we try replacing the ResNet50 classifier with a smaller but worse VGG16~\cite{VGG:Simonyan2014} classifier, with an average precision of $96.5$ (as opposed to the ResNet50's $97.5$).
This means the segmentation performance does not improve as much as it does with a ResNet50, but the difference is not too much for any models we experiment with (DeeplabV1~\cite{DeeplabV1:2015Chen}, DeeplabV3~\cite{DeeplabV3:2017Chen} and CCT~\cite{CCT:2020Ouali}).
In \cref{tbl:sota_with_vgg}, we show the segmentation performance using a VGG16 classifier along with a subset of results from \cref{tbl:sota}.
VGG in ``Pred. Filter'' column refers to prediction filtering with a VGG16 classifier, while ResNet refers to prediction filtering using a ResNet50 classifier.
Prediction filtering with a VGG16 classifier improves DeeplabV1 by a large margin ($61.8 \rightarrow 66.1$).  
It also improves DeeplabV3 by $1.4$, and CCT by $0.9$.

We also provide experimental results on various levels of supervision in \cref{fig:additional_various_supervision}, with and without a CRF.
We can observe that the prediction filtering using the VGG16 classifier indeed still significantly improves the mIoU, regardless of the number of pixel labels.
Full numbers for mIoU are provided in \cref{tbl:additional_various_supervision}.
The improvement is larger with DeeplabV1 and V3 than CCT;
recall that CCT already exploits some of the information from image-level labels via extracted pseudo-labels.

\begin{figure*}[t!]
  \centering
  \includegraphics[width=0.9\textwidth]{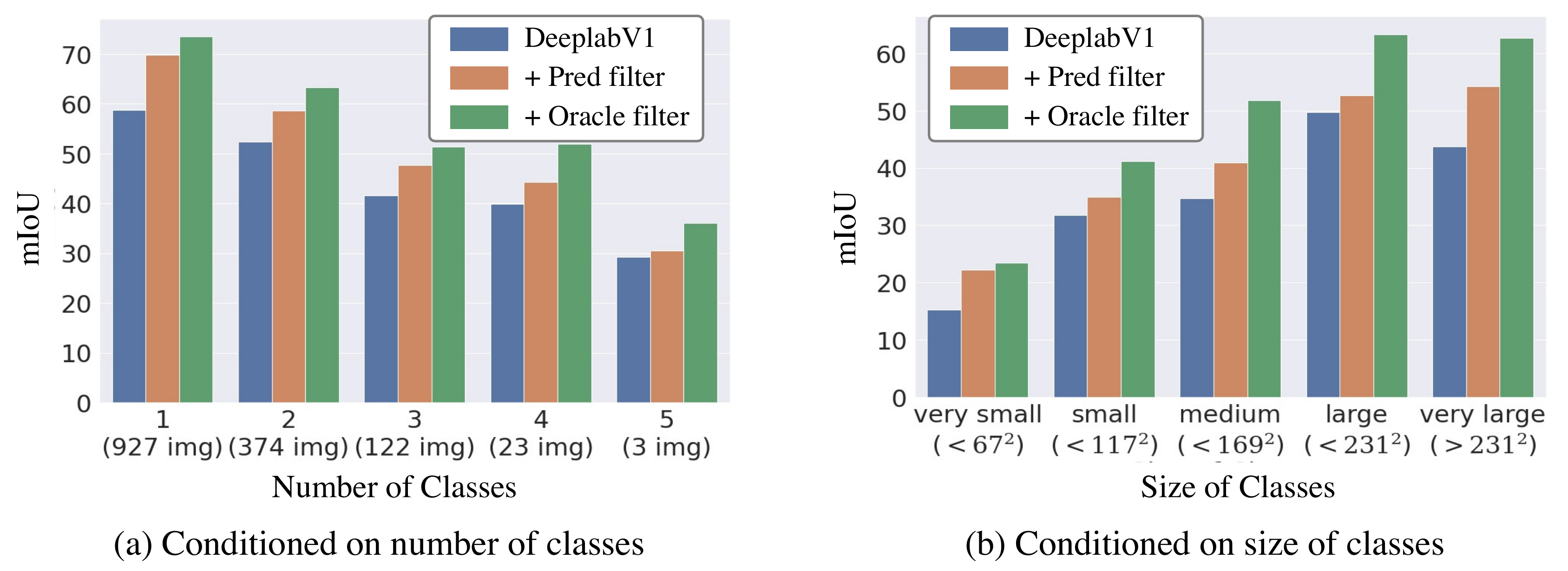}
  \caption{mIoU computed with and without prediction filtering only on certain sub-groups of images.}
  \label{fig:subgroup}
\end{figure*}

\section{Interaction of Prediction Filtering with the Global Context} 
\label{sec:global_context}

\begin{table}[h!]
\centering
\fontsize{9.0}{12.0}\selectfont
\begin{tabular}{cc|c|c|c} 
\hline
\multirow{2}{*}{CRF}  & \multirow{2}{*}{Filtering}  & w/o Atrous   & w/ Atrous  & w/ Atrous \\
 & & w/  ASPP & w/o ASPP  & w/ ASPP \\
\hline\hline
     &     & 68.1 & 63.3  & 70.6 \\
\cm  &     & 69.8 & 65.3 & 72.4 \\
\cm  & VGG & 71.3 (+1.5) & 69.2 (+3.9) & 73.8 (+1.4) \\
\cm  & ResNet & 72.6 (+2.8) & 70.3 (+5.0) & 74.9 (+2.5) \\
  \hline
\Xhline{2\arrayrulewidth}
\end{tabular}
\caption{Effect of the global context to prediction filtering.}
\vspace*{-2mm}
\label{tbl:global_context}
\end{table}

The success of prediction filtering comes largely from a classifier ``looking at images differently'' compared to a segmentation network. As we empirically demonstrate in \cref{fig:problems}, the classifier mainly captures the global features instead of local features as done by the segmentation network.
One nature question is then how the segmentation networks’s ability to capture the global context would affect the performance of prediction filtering.
It is an important question but is somewhat difficult to get at experimentally.
In an attempt to control the “level of global context” of DeeplabV3, we reverted some of the modifications made in its development~\cite{DeeplabV3:2017Chen} made with the goal of adding more context.
Specifically, we changed atrous convolutional layers of the last block of the ResNet backbone to regular convolutions by decreasing the dilation rate (“w/o Atrous” below), and removed atrous spatial pyramid pooling (ASPP; “w/o ASPP” below).

By decreasing the dilation rate, the output stride increased from $16$ to $32$ (for more detailed information, see Figure 5 of \cite{DeeplabV3:2017Chen}). Also, although the motivation of ASPP was to provide multi-scale features, it also does provide more context information by considering larger receptive fields in various scales. \cref{tbl:global_context} shows how each of the atrous conv layers and ASPP affect the segmentation performance with and without prediction filtering (using both VGG16 and ResNet50 classifier).

With ASPP present, Atrous does not dramatically change the amount of improvement due to prediction filtering. On the other hand, prediction filtering helps much more when ASPP is not present. This experiment certainly does not fully explain the relationship between a segmentation network's ability to capture global context and performance improvement due to prediction filtering – in particular, the case where prediction filtering helps more is also the case where baseline performance is much lower – but it does suggest that successfully incorporating more global context into a segmentation network may decrease how much prediction filtering helps.

\section{Alternative Approaches}
\label{sec:alternatives}
Perhaps the most obvious alternative to prediction filtering is a ``soft'' filter, one that, say, multiplies segmentation probabilities by classifier probabilities before choosing a prediction.

Doing this as a post-processing step is significantly better than the baseline performance but not as good as prediction filtering (baseline: $61.8$, prediction filtering: $67.4$, and soft filtering: $65.1$).
To understand the difference between soft and hard (prediction) filtering better, we reformulate classifier probabilities with temperature $T$ and a shift bias $\theta$ as follows,
    $$
        p(x) = \frac{1}{1 + \exp(-T(x - \theta))}
    $$
where $x$ is the output logit of the classifier.
For various $T \in [1, 2, 4, 10, 100, \infty]$, we find the optimal $\theta$ that gives the highest mIoU, and report the mIoU.
Note that $T=1$ is equivalent to soft filtering whereas $T=\infty$ is equivalent to prediction filtering.
The segmentation performance improves as $T$ increases, and is maximized when $T=\infty$.
\begin{table}[h!]
    \vspace*{3mm}
    \centering
    \fontsize{9.0}{12.0}\selectfont
    \begin{tabular}{c|cccccc}
    \hline
    T & 1 & 2 & 4 & 10 & 100 & $\infty$ \\
    \hline
    mIoU & 65.1 & 66.0 & 66.3 & 66.7 & 67.0 & 67.4  \\
    \Xhline{2\arrayrulewidth}
    \end{tabular}
    \caption{Variants of soft filtering.}
    \label{tbl:sigmoid}
\end{table}
If we do it at training time instead -- training the segmentation model with a frozen classifier from the start -- the model can adapt better, and in fact, it is compatible to prediction filtering ($67.0$ \vs $67.4$).
But, as it requires an additional training step, it is not as efficient as prediction filtering.

Since joint training helps with soft filtering, it might also help to train the segmentation model using standard prediction filtering.
Indeed, it does ($61.8 \rightarrow 66.5$), but not quite as much as joint training with soft filtering ($67.0$) or prediction filtering at inference time ($67.4$).
One possible explanation is that, while training with prediction filtering, the segmentation model does not get the opportunity to learn that, \eg the body pixels in \cref{fig:problems} are actually a \labelname{cat} and not a \labelname{horse}.
The loss becomes
\[
    -\sum_{i=1}^N \sum_{m=1}^M \log \frac{%
    \exp f(x_i)_{m,y_{i,m}} \mathbbm{1}[g(x_i)_{y_{i,m}} \ge \tau)]
    }{%
     \sum_{c=1}^K \exp f(x_i)_{m,c} \mathbbm{1}[g(x_i)_{c} \ge \tau)]
    }
;\]
thus, in this situation, gradient updates ignore filtered classes (such as \labelname{horse}) entirely,
substantially reducing the amount of supervision provided to the segmentation model during training.

Another alternative to consider is to share features between the segmentation and classification networks, as it would help to decrease the total number of parameters used. 
However, our attempts at this significantly degraded performance.
Consider training on only $200$ pixel-labeled images; this is probably where it is most likely feature sharing would help.
If the networks share the first three blocks of VGG16, the performance drops from $58.8$ to $55.6$.
The pattern is similar for other levels of supervision.
For prediction filtering to work, it is important that each network be able to ``do its job'' in its own way.
(Pseudo-label methods usually also use a separate classifier.)


\newcolumntype{C}{>{\centering\arraybackslash}p{2.6em}}
\begin{table*}[h!]
\centering
\fontsize{9.0}{12.0}\selectfont
\begin{tabu}{c|cccccc|C}
\hline
\multirow{2}{*}{Method} & \multicolumn{6}{c}{Hyperparameters} \\
\cline{2-8}
 & Optimizer & LR & Weight Decay & Nesterov & Grad. Ratio & Batch Size & $\tau$ \\
\hline
\hline
DeeplabV1~\cite{DeeplabV1:2015Chen} & SGD & 0.0010 & 0.0003 & False & 0.3 & 10 & -0.5 \\
DeeplabV3~\cite{DeeplabV3:2017Chen} & SGD & 0.0010 & 0.0003 & True & 0.1 & 8 & -0.5 \\
WSSL~\cite{WSSL:2015Papandreou} & SGD & 0.0010 & 0.0005 & False & 1.0 & 15 & -0.5 \\
CCT~\cite{CCT:2020Ouali} & SGD &  0.0100 & 0.0001 & False & 0.1 & 10 & -1.0 \\
DeeplabV1 \& 3-W~\cite{Dual:2020Luo} & Adam & 0.0001 & 0.0001 & False & 1.0 & 4 & -2.0 \\
DualNet (V \& R)~\cite{Dual:2020Luo} & Adam &  0.0001 & 0.0001 & False & 1.0 & 4 & -2.0 \\
AdvCAM~\cite{AdvCAM:2021Lee} & SGD &  0.0100 & 0.0001 & False & 0.1 & 16 & -1.0 \\
\Xhline{2\arrayrulewidth}
\end{tabu}
\caption{The hyperparameters used for different models.}
\vspace*{-1mm}
\label{tbl:hyperparam}
\end{table*}



\section{Performance on Sub-groups}
\label{sec:subgroup}

In Figure~\ref{fig:subgroup}, we compare the improvement of prediction filtering on different sub-groups; Figure~\ref{fig:subgroup}(a) and (b) are based on a group of images conditioned on the number of classes and size of classes, respectively.
Figure~\ref{fig:subgroup}(a) shows that while prediction filtering indeed works the best on single-class images, it still helps significantly for images with multiple classes present.
On the other hand, Figure~\ref{fig:subgroup}(b) shows the improvement in mIoU computed only for images whose number of pixels in the given class is in a particular size bracket. (As we only have semantic information, not instance information, object size cannot be obtained.)
We do not think there is a strong correlation between object size and improvement in mIoU of prediction filtering; the largest improvements are made in very small and very large.

\section{Sensitivity to Thresholds}
\label{sec:sensitivity}
\begin{figure}[t!]
  \begin{center}
    \includegraphics[width=0.46\textwidth]{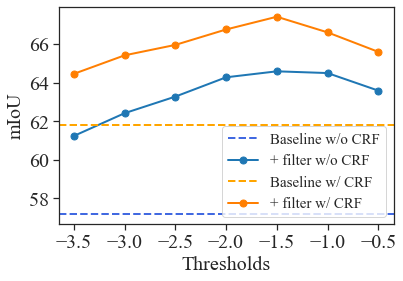}
  \end{center}
  \vspace*{-1.3em}
  \caption{The performance for different filtering thresholds.}
  \label{fig:thresholds}
  \vspace*{-1.2em}
\end{figure}

We have also performed the experiments showing how prediction filtering threshold $\tau$ affects the segmentation performance.
Figure~\ref{fig:thresholds} shows the segmentation performance for different filtering thresholds (in x-axis) with and without CRFs applied.
As shown in the figure, the performance is not very sensitive to the choice of threshold in the reasonable wide range ($-3.5 \leq \tau \leq -0.5$).

\section{Reproducibility}
\label{sec:reproducibility}
As shown in the main paper, we provide the experiments with DeeplabV1~\cite{DeeplabV1:2015Chen}, DeeplabV3~\cite{DeeplabV3:2017Chen}, WSSL~\cite{WSSL:2015Papandreou}, CCT~\cite{CCT:2020Ouali}, DualNet~\cite{Dual:2020Luo}, and AdvCAM~\cite{AdvCAM:2021Lee}.
We re-implement DeeplabV1 largely based on the  Pytorch implementation (\httpsurl{github.com/wangleihitcs/DeepLab-V1-PyTorch}).
For the other models, we apply the oracle and prediction filtering on the original implementation of the publicly available code -- DeeplabV3 (\httpsurl{github.com/chenxi116/DeepLabv3.pytorch}), WSSL (\httpsurl{github.com/TheLegendAli/DeepLab-Context}), CCT (\httpsurl{github.com/yassouali/CCT}), and AdvCAM (\httpsurl{github.com/jbeomlee93/AdvCAM}), and the code obtained from the original authors of DualNet.
The implementation of DeeplabV3, CCT, DualNet and AdvCAM is based on Pytorch, while WSSL is implemented using Caffe.
We provide an anonymous and reproducible code for the DeeplabV1 and proposed prediction filtering method in zip file along with this appendix.

\paragraph{Hyperparameters.}
In \cref{tbl:hyperparam}, we provide the hyperparameters used for the experiments provided in the main paper.
Recall that $\tau$ is the prediction filtering threshold that determines the presence of a class given the output logits from a classifier.
For the Deeplab baselines, we tune the hyperparameters to boost up the performance on various levels of supervision.
On the other hand, for WSSL, CCT, AdvCAM and DualNet, we use the same hyperparameters provided in the original implementation.
The same hyperparameters are used with and without prediction filtering, as mentioned in the main paper.
Also, we follow previous practice in training DeeplabV1 for $20$ epochs and DeeplabV3 for $50$ epochs.



\begin{figure*}[t!]
    \centering
    \includegraphics[width=0.93\textwidth]{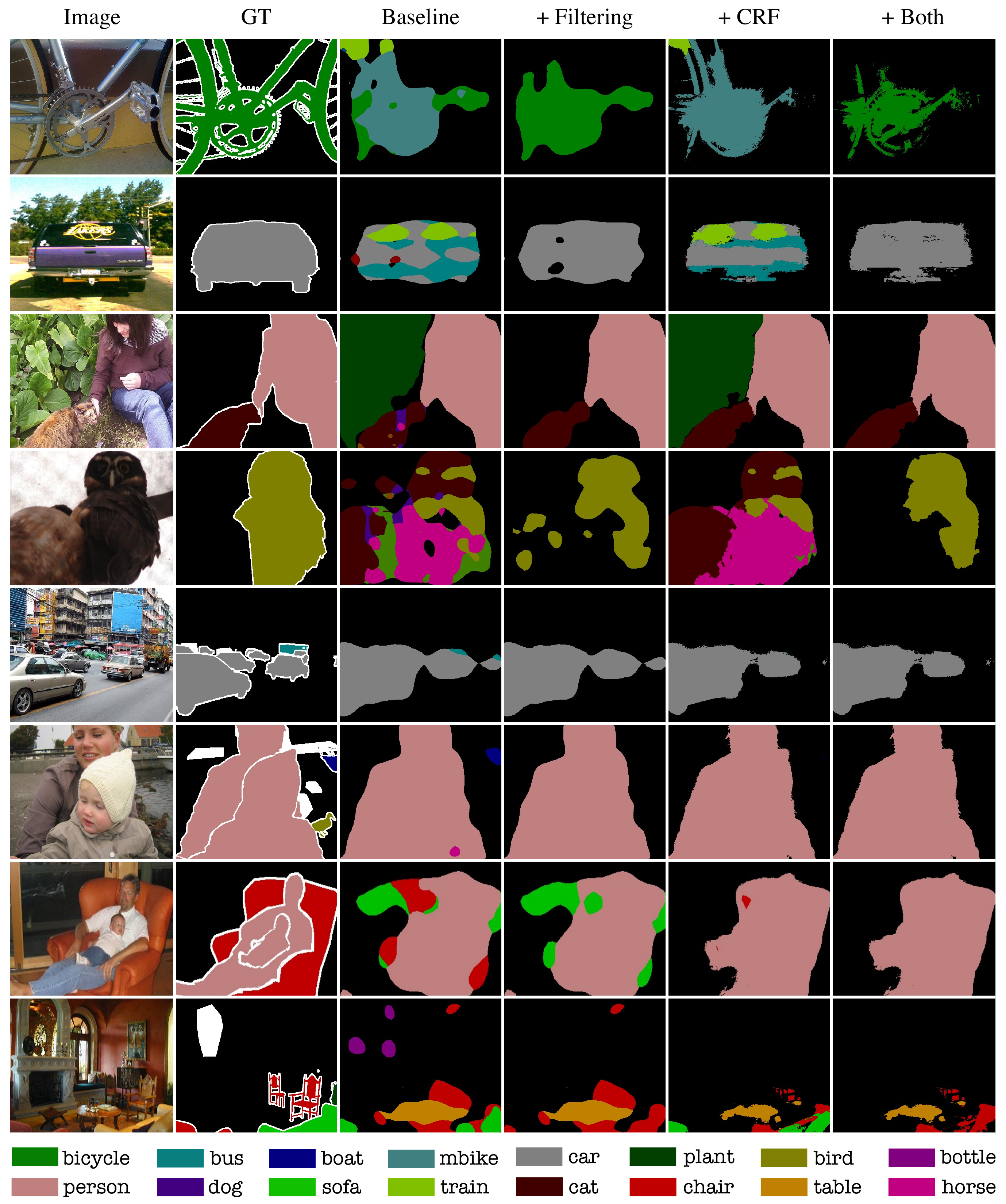}
    \vspace*{-1mm}
    \caption{Qualitative results. First and last four rows show the success and failure cases, respectively.}
    \label{fig:additional_qualitative}
    \vspace*{-1mm}
\end{figure*}

\section{Additional Qualitative Results}
\label{sec:additional_qualitative}
Lastly, \cref{fig:additional_qualitative} gives additional qualitative results using DeeplabV1 trained on $1{,}464$ images with pixel-level labels.
The first half of the images are successes of prediction filtering, and the last half failures.

As shown in the first two images, the segmentation network is often confused between similar classes, \eg \labelname{bicycle} and \labelname{mbike} (motor bike), or \labelname{car} and \labelname{bus}.
Since the area of wrong prediction is fairly large, CRF actually makes the prediction worse.
The prediction filtering, however, safely removes the wrong predictions.
The examples in the third and fourth rows show cases where the segmentation network incorrectly predicts the background region into some object classes.
As the classification network does not detect any of the irrelevant classes predicted by the segmentation network, the prediction filtering corrects the predictions in a large region of an image.

As discussed in the main paper, however, the method is not perfect.
In cases where objects are too small or occluded, the classifier may predicts that the corresponding class is not present.
For examples, in the fifth and sixth image, it is not easy to identify a \labelname{bus} and \labelname{bird}, since they are too small.
Or, for the cases like the seventh and eighth images, where the \labelname{chair} and \labelname{sofa} are largely occluded, the classifier is sometimes confused.
(In the seventh image, the \labelname{chair}/\labelname{sofa} distinction is arguably confusing even for humans.)

\end{document}